\documentclass[10pt,twocolumn,letterpaper]{article}

\usepackage{wacv}
\usepackage{times}
\usepackage{epsfig}
\usepackage{graphicx}
\usepackage{amsmath}
\usepackage{amssymb}
\usepackage{color, soul}
\usepackage{multirow}
\usepackage{array}
\usepackage{enumitem}
\usepackage{authblk}

\definecolor{darkolivegreen}{rgb}{0.33, 0.42, 0.18}
\definecolor{matt_color}{rgb}{0.0, 0.42, 0.24}
\newcommand{\rev}[1]{{\color{black} {#1}}}

\def\wacvPaperID{247} % Enter the WACV Paper ID here

\wacvfinalcopy % *** Uncomment this line for the final submission

\ifwacvfinal
\usepackage[breaklinks=true,bookmarks=false]{hyperref}
\else
\usepackage[pagebackref=true,breaklinks=true,colorlinks,bookmarks=false]{hyperref}
\fi

\begin{document}

\newcolumntype{?}{!{\vrule width 1pt}}
%%%%%%%%% TITLE
\title{Single Image Human Proxemics Estimation for Visual Social Distancing}

\author[1]{Maya Aghaei} 
\author[1,2]{Matteo Bustreo}
\author[1]{Yiming Wang}
\author[1]{Gianluca Bailo}
\author[1]{Pietro Morerio} 
\author[1]{\\Alessio {Del Bue}}
\affil[1]{Pattern Analysis \& Computer Vision (PAVIS), Istituto Italiano di Tecnologia, Genova, Italy}
\affil[2]{Dipartimento di Ingegneria Navale, Elettrica, Elettronica e delle Telecomunicazioni, University of Genova, Italy}

%{\tt\small \{name.surname\}@iit.it}\\

\maketitle
%\thispagestyle{empty}

%%%%%%%%% ABSTRACT
\begin{abstract}
In this work, we address the problem of estimating the so-called ``Social Distancing'' given a single uncalibrated image in unconstrained scenarios. Our approach proposes a semi-automatic solution to approximate the homography matrix between the scene ground and image plane. With the estimated homography, we then leverage an off-the-shelf pose detector to detect body poses on the image and to reason upon their inter-personal distances using the length of their body-parts. Inter-personal distances are further locally inspected to detect possible violations of the social distancing rules. We validate our proposed method quantitatively and qualitatively against baselines on \rev{public domain} dataset\rev{s} for which we provided groundtruth on inter-personal distances. Besides, we demonstrate the application of our method deployed in \rev{a} real testing scenario where statistics on the inter-personal distances are currently used to improve the safety in a critical environment. 
\end{abstract}

%%%%%%%%% BODY TEXT
\section{Introduction}
\label{sec:intro}

The recent worldwide pandemic emergency raised attention on daily-life circumstances which previously were not a cause of concern. Among them, there are constraints on the physical distance between people as an effective measure to reduce the virus spread. However, beyond the enforcement of such rules, a critical issue for safety is to verify and quantify the actual compliance of people with these restrictions which indeed have a substantial impact on our social life \cite{brooks2020psychological}. \rev{To this end, a lot of solutions have been proposed \cite{johnson2020social}. However, cameras in video surveillance settings offer arguably a more viable infrastructure to control the so-called \textit{Social Distancing} (SD).}

\textit{Visual Social Distancing} (VSD) \cite{vsd2020} is a particular case of the SD estimation problem, in which the inter-personal distance is estimated from a single uncalibrated image or video. VSD solutions use camera networks that are often deployed in pre-existing video surveillance settings. This allows fast integration of VSD to increase the safety of the population, by detecting recurrent SD violations or by generating statistic analysis. This information can be used to identify risky areas that are subject to crowding and to redefine the architectural design to improve safety in public and private places. 

\begin{figure}
    \centering
    \includegraphics[width=1.\columnwidth]{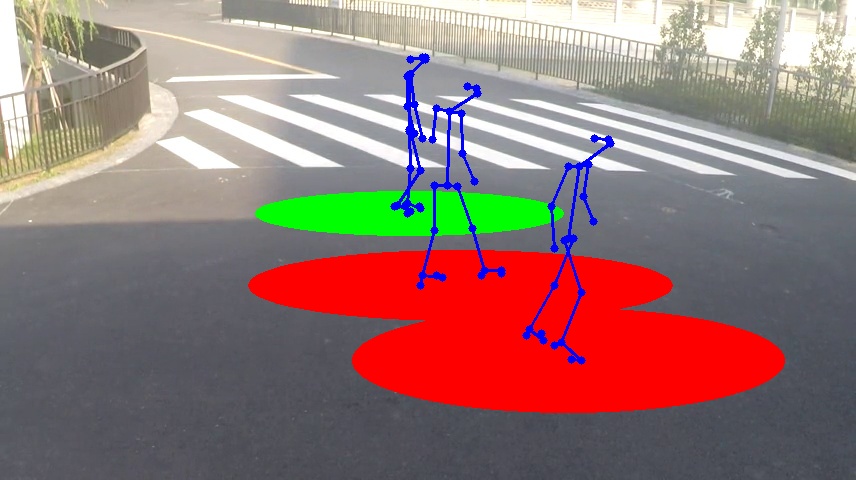}
    \caption{An example of the output of our proposed pipeline. The pedestrians are modeled as skeletal figures and each disc on the ground plane represents the minimum allowed inter-personal distance. The violation of social distancing is shown in red.}
    \label{fig:teaser}
\end{figure}

The study in \cite{vsd2020}, although inspiring, does not propose an exact methodology for solving VSD, but rather provides a detailed overview of general issues to be addressed as well as a few potential guidelines to implement solutions. Differently we propose a practical method that estimates a safe space for each detected person \rev{(i.e. person proxemics)} on uncalibrated images, and detects violations of SD rules by measuring overlapping of safe spaces across people in the image scene (see Fig. \ref{fig:teaser}). 

Estimating VSD requires the computation of metric inter-personal distances (e.g. $\leq 2m$) between detected people in the scene. A convenient region to estimate distances is the ground plane where people stand, strengthened by the fact that locally it can be considered as planar. We can therefore estimate a local planar homography between the ground plane and the image plane and relate them to a known metric reference.

\rev{Automatic homography estimation or camera calibration previously appeared as stand-alone problems in the computer vision literature \cite{Hartley2004, szeliski2011}. However, to the best of our knowledge, there is no prior work that proposes a full solution for inter-personal distance estimation from a single uncalibrated image. Our method is a practical solution that takes advantage of the common installation scenario of surveillance cameras to perform both homography estimation and \textit{local} metric inference. We exploit the standard surveillance settings, where people appear up-right, relying only the fact that surveillance cameras are typically placed at a certain height with a certain tilt angle from the ground which leads to approximately a uniform projection of lengths across cameras.
We estimate the homography matrix between the bird's eye view ground plane and image plane by manually selecting two intuitive auxiliary perspective ratios. Upon estimation of homography, people are detected and a local reference is inferred from them to estimate pairwise distances.  With subjective evaluation, we demonstrate that auxiliary ratio parameters, even selected by human users, offer close proximity with the groundtruth values obtained from the actual homography. Moreover, we choose to use and validate body pose detectors since they offer richer information about the 2D body joints, whose length and relative position can be leveraged to provide a metric reference.\footnote{Note that human body detection is not the focus of this work, and we rely on off-the-shelf methods (e.g. OpenPose \cite{openpose:ijcv}).}
}

% %
% %sufficiently stop an average person from bashing it, while maintaining the camera visibility for additional security purposes. 
% Given such camera constraints, for any given image with people detected, we estimate an homography matrix between the bird's eye view ground plane and perspective image plane by manually selecting two intuitive auxiliary perspective ratios. % to support the manual selection of key correspondence points. 
%Another important yet indirect advantage of using a pose detector is for privacy preserving purposes. The estimated joints of a human body allows transferring the analysis of interpersonal distance measures from the actual image of an human to their skeleton in the scene, while preserving the necessary information for VSD evaluations (see the joints plus VSD disc in Fig. \ref{fig:teaser}). 

%Our method is evaluated on the first real-world VSD dataset against baselines in order to validate the design choices of the approach. Additionally, as the proposed framework is currently being tested in critical infrastructures, we show statistics over its 24/7 output that is being used at this moment to facilitate faster and better adaptation of spaces to new SD regulations. 

\rev{We have evaluated our method on VSD-adapted public datasets against baselines in order to validate the design choices of the approach. Additionally, as the proposed framework is currently being deployed in real-world environments, we show real statistics given by the system being deployed 24/7 in one of these scenarios to facilitate faster and better adaptation of spaces to new SD regulations.} The main contributions of this work are:
\begin{itemize}[noitemsep,topsep=0pt]
    \item We propose the first\footnote{At the time of submission there were no previous peer-reviewed works addressing VSD.} VSD method to identify the \rev{local human proxemics} for automatic SD rule violation detection on single uncalibrated images.
    \item We propose a semi-automatic approach for homography approximation between ground plane and image plane by intuitively selecting two auxiliary ratios.
    \item \rev{We provide an analysis over human body joints to select the best body parts to be used as reference for scene metric inference.}
    \item \rev{We validate various stages of our method over three public datasets that we labeled with pairwise human distance measures to adapt them for VSD analysis. Code is available at \url{https://github.com/IIT-PAVIS/Social-Distancing}.}
    \item We present the advantages and challenges of deploying our proposed method deployed in real scenarios.
\end{itemize}

%-------------------------------------------------------------------------

\section{Related work}
\label{sec:relawork}

\rev{According to \cite{vsd2020},} estimating VSD, requires the solution of a set of computer vision tasks, namely, scene geometry understanding and person detection/body pose estimation. Indeed, the geometry of the scene is the first important step to define a local metric reference for measuring inter-personal distances. The second yet equally important task, is the detection of people in the scene in a possibly crowded environment. Once the target people are correctly localised in the scene, their distance can be locally estimated to understand if their mutual distance is lower than the predefined threshold (\eg $\sim2m$). \rev{In this work, our focus is on scene geometry understanding from monocular images, instead of using camera calibration, or other sensory inputs such as stereo images \cite{pseudo_lidar++} or depth \cite{detection_depth}. In this regard, the most similar work to us is Monoloco \cite{bertoni2019monoloco}, in which authors propose a solution for 3D human localization from a monocular camera. The model behind Monoloco is a feed-forward neural network that predicts human 3D localization. Eventually, the VSD can be calculated from the 3D position of the people in the scene. Despite similarities between Monoloco and our proposed algorithm in simplicity and objectives, in contrast to Monoloco our proposed model does not require any camera calibration information (not even a default value) nor a training procedure, and it does not intend to solve the VSD problem globally through 3D localization of pedestrians but locally for each pair of people.} In the following, as there is no further previous work on estimating VSD from a single uncalibrated image, we will describe the related work to the two main sub-tasks involved in the solutions for this problem.

\begin{figure*}
    \centering
    \includegraphics[width=2\columnwidth]{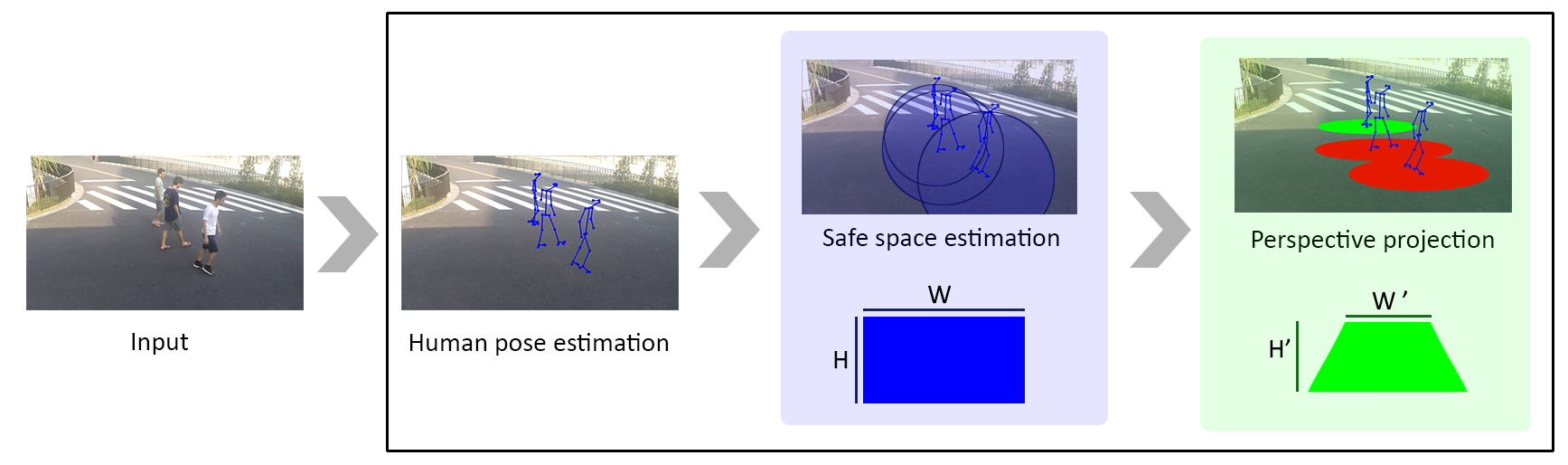}
    \caption{A schematic representation of our proposed pipeline for estimation of safe space and detection of its violation, comprised of three main computational stages: body-joint detection, body height estimation for metric inference, and ground-plane geometry estimation.}
    \label{fig:method_scheme}
\end{figure*}
\textbf{Scene Geometry Understanding}
% \label{sec:local_geometry}
The task of measuring social distancing from images requires the definition of a metric reference. This problem  is strongly related to the single view metrology topic \cite{criminisi2000single} as we consider the most common case of a fixed camera. An initial solution for estimating inter-personal distances requires the identification of the ground plane where people stand \cite{lee2000monitoring,hoiem2007recovering,groundnet:2019,Abbas:az:2019,watson-2020-footprints}. Such ground plane serves in many video surveillance systems to visualise the scene as a bird's eye view for ease of statistics representation. Most works impose the assumption that the ground plane is planar. Then, the problem is to estimate a homography given some reference elements (\eg, known objects or manual measurements) extracted from the scene or using the information of detected vanishing points in the image \cite{criminisi2000single, rother2002new, mirzaei2011optimal, wildenauer2012robust, bazin2012globally, bazin20123, zhang2016vanishing, lu20172}. Another common approach is to calibrate fixed cameras by observing the motion of dynamic objects such as pedestrians \cite{liu2011surveillance,lv2006camera,tang2019esther,xu2020estimating}. As a unique approach, AutoRect \cite{chaudhury2014auto} targets the restoration of bird's view homography from the perspective scene image by restoring the parallelism of lines by computing 2D homography from only two estimated vanishing points, allowing the algorithm to be used in more general scenarios. Furthermore, approaches based on deep learning also attempt at estimating directly camera pose and intrinsic parameters on a single image \cite{hold2018perceptual, Lopez_CVPR_2019}. \rev{In this work instead, we propose a simple yet effective approach that does not require exhaustive training procedures but only two auxiliary perspective ratios for approximating the homography matrix.}   

Nonetheless, VSD estimation in addition to camera intrinsic/extrinsic parameters estimation or ground plane detection requires a metric reference. Such information can be coarsely computed in the scene given objects of known dimension or by using a standardized height of pedestrians as a rule of thumb \cite{wagner2019, benabdelkader2008statistical, vester2012estimating}. Similarly, in this work, we rely on a set of human body joints to infer the metric reference. \rev{It is important to note that a local solution for VSD is a simpler problem than global solutions which entail estimation of every metric distance among people at any position in the image. While the global solutions can be hard to estimate, for example, when the single ground plane assumption is violated, local solutions only require distance measure calculation when two or more pedestrians get close enough for triggering the necessity of a measure.} 

\textbf{Person Detection and Pose Estimation}
% \label{sec:person_detection}
Person detection has reached impressive performance in the last decade given the interest in the automotive industry and other application fields \cite{benenson2014ten}. Real-time approaches can now estimate human pose in complex scenarios \cite{openpose:ijcv} and even reconstruct a 3D mesh of the person body \cite{guler2018densepose}. The majority of the approaches estimate not only people's location as a \rev{bounding box} but also 2D stick-like figures, thus conveying a schematic representation of the pose. Recently, several methods augment 2D poses in 3D or infer directly a 3D pose in a normalised reference system \cite{wang2014robust,moreno20173d,tome2017lifting,mehta2017vnect,martinez2017simple, bogo2016keep, mehta2019xnect,rogez2019lcr,zhou2017towards}. 

%An example of a complex scenario and the robust performance of \cite{openpose:ijcv} is shown in Fig. \ref{fig:vitruvian}.

Specific pedestrian detection techniques have been designed to work in crowded scenes \cite{vandoni2019evidential,Wang_2018_CVPR,Liu_2019_CVPR,ge2020ps}, where skeleton-based representations are often dropped in favor of saliency-based masks, especially focusing on heads. When the image resolution becomes too low to spot single people, regression-based approaches are employed \cite{boominathan2016crowdnet,liu2018decidenet,setti2018count,padnet2020}, providing in some cases density measures \cite{sindagi2018survey,rangel2017entropy,sindagi2017cnn}. This information, merged with a geometric model of the scene, can lead to a solution for measuring the average SD in the field of view. Recently, new efforts tend towards solving body detection and pose estimation in crowded environments \cite{golda2019human,li2019crowdpose}, the very same scenario that SD is trying to impede. Yet, finding the location of people in such cases is of relevant importance for buzzing alerts or creating statistics of overcrowded areas. To this end, the people detection module has to be robust to severe self and other occlusions, different image scales, and indoor/outdoor scenarios. 

%\begin{figure}
%    \centering
%    \includegraphics[width=0.8\columnwidth]{./pictures/vitruvian-openpose.jpg}
%    \caption{OpenPose \cite{openpose:ijcv} as our selected pose detector performance over \textit{Vitruvian Man} Roman painting by \textit{Leonardo da Vinci}.}
%    \label{fig:vitruvian}
%\end{figure}

It is worth to highlight that, person detection in \rev{bounding box} format does not account for different body poses (\eg, sitting, riding) and might negatively impact the estimation of height and thus cause wrong SD analysis. In this case, finding joints and body parts has certain advantages. This is due to the fact that for obtaining an approximate of the metric reference, or even calibrating the cameras, usually the person height is used as a coarse proxy as computed from a \rev{bounding box} or using more precise techniques \cite{wagner2019, vester2012estimating,dey2014estimating,Gunel_2019_ICCV}. Given a metric reference from scene geometry and detected people pose in the scene, the VSD can be solved as a distance on the ground plane among the detected pedestrians. As previously discussed, this information can be estimated locally or pairwise in order to reduce the complexity of estimating a global reference system for the whole image.

%-------------------------------------------------------------------------

\section{Method}
\label{sec:method}

The goal of our proposal is to detect SD violations by identifying the overlapping of the circular safe space around the pedestrians using images from any uncalibrated cameras. To achieve this goal, our pipeline is comprised of three main steps: the ground-plane homography estimation, person detection and localisation, and joint-based metric inference (see Fig. \ref{fig:method_scheme}). Given any image captured by an uncalibrated camera, we first estimate the homography matrix between the ground plane and image plane (up to a scale) in a semi-automatic manner. We then detect people in the image using a human body-joint detector, e.g. OpenPose \cite{openpose:ijcv}, to exploit the rich information embedded in the human pose. Finally, for each detected person, we analyse the body joints to robustly localise its ground-plane centroid on the image, and then to infer the safe space and detect the SD violation. In the following, we detail each of the three modules of our proposed method.

\subsection{Homography Estimation}

\label{sec:homography}
\begin{figure}
    \centering
    \includegraphics[width=\columnwidth]{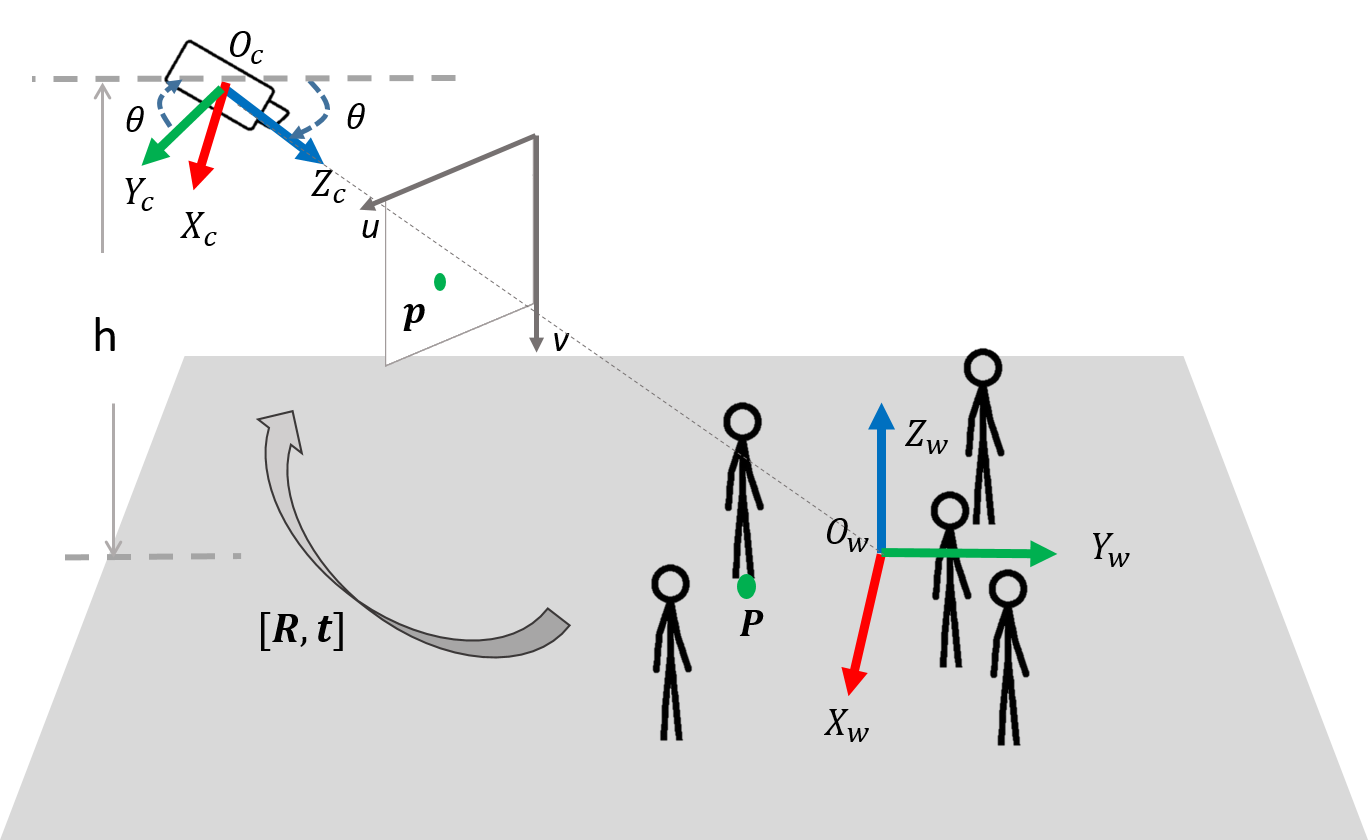}
    \caption{\rev{Illustration of the world and camera coordinate system under our assumptions that i) the camera only has a tilt angle $\theta$ with zero roll and pan angle, and ii) the camera has the height $h$ from the ground and the world origin locates at where the camera principal axis intersect on the ground plane. The $\bold{P}$ on the ground plane can be mapped to the pixel $\bold{p}$ on the image through the homography matrix.}}
    \label{fig:camera_setup}
\end{figure}

In this section, we cover the approach to estimate the homography matrix of any given image plane to the ground plane. Let us consider a pinhole camera model with the camera intrinsic matrix $\mathbf{K}\in \mathbb{R}^{3\times3}$
% %= \begin{bmatrix}
% f_x& 0 & c_x\\ 
% 0 & f_y & c_y\\ 
% 0 & 0 & 1
% \end{bmatrix}$, 
and extrinsic matrix $[\bold{R}|\bold{t}]$. The intrinsic matrix $\bold{K}$ contains the camera parameters that are related to the focal length, skew, and the principal point. The extrinsic matrix $[\bold{R}|\bold{t}]$ denotes the 3D coordinate transform from the world to the camera reference, where $\bold{R} \in SO(3) $
is the rotation matrix and $\bold{t} \in \mathbb{R}^{3\times1}$ is the translation vector. 
Any 3D point in the world coordinate $\bold{P}$ can be projected to the image plane at pixel position $\bold{p}$ as:

\begin{align}
\centering
\label{eq:projectionM}
s\bold{p} =\bold{K}[\bold{R}|\bold{t}]\bold{P} = \bold{K}\begin{bmatrix}
\bold{r}_{1} & \bold{r}_{2} & \bold{r}_{3} & \bold{t}
\end{bmatrix}\bold{P},
\end{align}
where $s$ is the scale factor and both $\bold{P}=\left[X, Y, Z, 1\right]^T$ and $\bold{p} = \left[u, v, 1\right]^T$ are in its homogeneous coordinate format. 

In the case of projecting 3D points onto the ground plane, i.e. $Z=0$, to the image plane, Eq.~\ref{eq:projectionM} can be simplified as:
\begin{equation}
\centering
s\bold{p} = \bold{K}\begin{bmatrix}
\bold{r}_{1} & \bold{r}_{2} & \bold{t}
\end{bmatrix}\bold{P}' = \bold{H}\bold{P}',
\label{eq:homography}
\end{equation}
where $\bold{H}$ is the {\em homography matrix} to project points on the ground plane with metrics to the image plane and $\bold{P}' = \left[X, Y, 1\right]^T$.

In our problem setting, both $\bold{K}$ and $[\bold{R}|\bold{t}]$ are not available for the computation of homography matrix $\bold{H}$, we therefore approximate $\bold{H}$ (up to a scale) with reasonable assumptions based on the standard scenario observed from surveillance cameras. One key observation is that people are appearing (almost) upright in the image plane, which means we can assume the roll angle to be $0^{\circ}$. \rev{Moreover, we can set the camera pan angle as $0^{\circ}$ and the world origin at the intersection point of the camera principal axis on the ground plane with the height as $h$ (as shown in Fig. \ref{fig:camera_setup}). Let $\theta$ be the camera tilt angle, we will have $\bold{r}_{1}=[1, 0, 0]^{T}$ and $\bold{r}_{2}=[0, -\cos(\theta), -\sin(\theta)]^{T}$ and $\bold{t}=[0, -\frac{h}{tan(\theta)}, h]^{T}$.} Under such assumptions, lines that are parallel to the X-axis in the ground plane remain horizontal in the projected image plane, and lines parallel to the Y-axis converge to the vanishing point along Y-axis in the image plane. Therefore, a rectangular area on the ground plane with a width $W$ and a height $H$ would be projected into an isosceles trapezoidal shape with short-based width $W^{'}$ and height $H^{'}$ (see Fig. \ref{fig:method_scheme}), where the horizontal ratio $\rho_{h} = \frac{W^{'}}{W}$ and vertical ratio $\rho_{v} = \frac{H^{'}}{H}$ are purely related to the camera tilt angle $\theta$ and the camera height $h$.

By manually tuning $\rho_{h}$ and $\rho_{v}$, one can locate the corresponding four corners between the rectangular shape and the projected isosceles trapezoidal shape, in order to estimate the homography matrix $\Tilde{\bold{H}}$ up to a scale. Note that $\Tilde{\bold{H}}$ does not reflect the real metric mapping between pixel and length as $\bold{K}$ is not known (for details regarding metric inference, please refer to Sec.~\ref{sec:pose-height}). Approximated homography $\Tilde{\bold{H}}$ however, can be used to project the circular safe space around each person, to form an approximation of an ellipse on the image plane in perspective. With subjective evaluation, we prove that human choice is reliable and can approach the optimal solution with grid search using the groundtruth inter-personal distances. Using these two ratios to facilitate the manual choice of the correspondence points, we can achieve a reasonable homography estimation when the camera calibration is not available.

\subsection{Image-Plane Person Detection and Localisation}
\label{sec:person_localisation}

We detect people on the image using a human body detector, e.g. OpenPose \cite{openpose:ijcv}, that provides the person detection as the set of detailed body joints. Let $\{\bold{D}_i\}_{i \in [1,N]}$ be the set of person detections, where $N$ denote the total number of detected people on the image. Each detection $\bold{D}_i$ is composed of a set of body joints, i.e. $\bold{D}_i=\left \{\bold{j}_i^{k}\right \}_{k\in\left[1,K\right]}$, and each joint corresponds to a position on the image plane $\bold{j}_i^{k} = [u_{i}^k, v_{i}^{k}]$, where $K$ is the number of joints that are supposed to be detected ($K=25$ for OpenPose). Out of the detected joints $\bold{D}_i$, we obtain the \rev{bounding box} $\bold{B}_{i}= \left[u_{i}^{min}, v_i^{min}, u_{i}^{max}, v_i^{max} \right]$ with $\left[u_{i}^{min}, v_i^{min}\right]$ being the minimum and $\left[ u_{i}^{max}, v_i^{max}\right]$ being the maximum among all detected joints. 

By localising a detected person, we mean to compute the pixel position $\bold{p}_i = [{u}_i, {v}_i]$ based on $\bold{D}_i$ that represents the person on the ground plane. It is necessary to guarantee that $\bold{p}_i$ locates on the ground plane as the homography matrix $\Tilde{\bold{H}}$ is estimated w.r.t it. To this end, we only consider person detections as valid if at least one joint of the feet is detected. For each valid detection, we decouple the estimation of ${u}_i$ and ${v}_i$ and make use of the rich information provided by the detected joints for a more robust localisation. 

Regarding ${u}_i$, we exploit the fact that, for a standing-up or sitting person, certain part of human body, i.e. \textit{head}, \textit{neck}, or \textit{torso} are often located on the middle line of the body. When any of the above mentioned joints are detected, we represent ${u}_i$ by their averaged horizontal pixel position. If none of the aforementioned parts are detected, ${u}_i$ is represented by the horizontal pixel position corresponding to the centroid of the \rev{bounding box} $\bold{B}_{i}$, i.e. ${u}_i = \frac{u_{i}^{min}+u_{i}^{max}}{2}$. 
For ${v}_i$, as the valid detection inevitably contains joints corresponding to feet, we can represent it as the vertical pixel position averaged over all the detected joints of feet. 

\subsection{Metric Inference}
\label{sec:pose-height}

%\adb{FIXME PIETRO/MAYA: this has to be solid, account for occlusions, a more robust way to compute height, use multiple people to have a mean, also how to transfer height into the planar distance???}

The specified \textit{safe space} around the detected person $\bold{D}_i$ in the scene, is a circular area centered at the person with radius $r$, which is set according to the SD rules. Theoretically, the mapping from the length unit to the pixel unit on the ground plane can be obtained by calibrating the camera. However, when the calibration is not available, metric inference from pixel becomes a challenging problem to tackle. We, therefore, propose a practical metric inference solution that can be applicable to many surveillance scenarios, using the structural priors of the human body.

Studies on the average human body height over the years demonstrated that human body height has only marginal differences (a few centimeters) among different races \cite{baten2014human}. In addition to body height, the length of body parts also maintain relatively stable attributes \cite{murtinho2015leonardo}. Thus, the human body potentially can be used to infer approximate length metrics for the image pixels. \rev{In this regard, given that we have estimated the homography between the ground plane and image plane, it would be a natural choice to use a set of human joints parallel to the ground plane, such as \textit{shoulders} or \textit{hips} for inferring the length metric. However, these joints are only reliable if the person appears perpendicular to the camera principal axis and any rotation of the body from this angle leads to a large variation onto the image plane.}

Other joints such as  \textit{legs},  \textit{arms}, or  \textit{torso}, are more robust in maintaining a proper metric reference, but they lie mostly on the vertical plane that is orthogonal to the ground plane. This means that such  metric references cannot be directly applied to measure distances on the ground plane. However, under most surveillance settings, where the camera is installed at a height and a tilt angle of a certain range, we can use the body parts on the vertical plane to practically approximate the metric on the ground plane. More specifically, onto the image plane, we consider that the safe space of a detected person is within a circle centered at $\bold{p}_i$ with a radius $r$ that can be set based on the pixel length of those vertical body parts. As validated in the experimental section, \textit{torso} proves to be the most robust option for metric reference estimation among different body parts. The circle is then projected with the homography matrix into an approximated ellipse shape. The violation of SD is detected if the ellipse of a detected person overlaps with any ellipse of other detected people, leading to an SD $\leq 2m$.

%https://healthyliving.azcentral.com/correlation-between-arm-length-leg-length-11071.html

%%%%%%%%%%%%%%%%%%%%%%%%%%%%%%%%
\section{Experiments}

\subsection{Evaluation Datasets}
%\paragraph*{Synthetic datasets:}

\begin{figure*}[ht!]
    \centering
    \includegraphics[width=2\columnwidth]{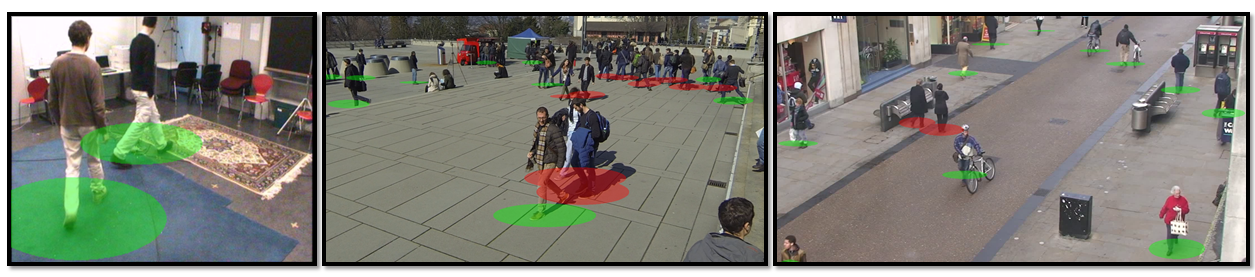}
    \caption{Image samples from the three datasets for the method evaluation with VSD results using the best parameters overlaid. Samples from left-to-right are from Epfl-Mpv-VSD, Epfl-Wildtrack-VSD, and  OxTown-VSD.}
    \label{fig:datasets}
\end{figure*}

For quantitative evaluation, we use real recordings of people in both indoor and outdoor scenes from  publicly available datasets. In these datasets, the 3D location of people on the ground plane can be estimated using either the homography matrix or the given camera intrinsic and extrinsic. In particular, we prepared three \rev{VSD-related}  datasets,  \textit{Epfl-Mpv-VSD}, \textit{Epfl-Wildtrack-VSD}, and \textit{OxTown-VSD}, \rev{out of the publicly available Epfl-Mpv \cite{Epfl-mpv2008},  Epfl-Wildtrack \cite{wildtrack2018}, and OxTown \cite{OxTown2011}, respectively.} In all these datasets, for each frame we compute the groundtruth as the pairwise inter-person distance across people using their 3D location in the scene as provided by the benchmark. 

We make use of the Epfl-MPV dataset as a dataset that covers an indoor environment. Epfl-MPV is originally used for multi-camera multi-person tracking, yet, our Epfl-Mpv-VSD uses only the lab sequences of Epfl-MPV, where there are 4 sequences recorded by 4 cameras that cover a lab area with up to 6 people moving freely inside. All sequences are synchronised and each sequence has 2955 frames. Epfl-Wildtrack-VSD is composed of seven synchronised sequences from the Epfl Wildtrack HD dataset served for multi-camera multi-people tracking. All sequences have a large overlapping field of view covering an open outdoor area, and each of them is composed of 401 frames. OxTown-VSD makes use of the OxTown sequence of 7498 frames, that covers a daily street in Oxford from a single static camera. Although the two datasets are featuring outdoor environments, the camera settings are similar to indoor surveillance cameras, therefore applicable to our method evaluation. 

We applied OpenPose \cite{openpose:ijcv} to all frames within the above-mentioned datasets to detect people in the format of 25 skeletal joints. A detection is considered to be valid only if at least one joint from the feet is detected and at least a total of 13 joints of the person are detected. For each frame, the groundtruth pairwise inter-person distance is computed among valid detections using the homography matrix. For both Epfl-Wildtrack-VSD and OxTown-VSD, we make use of the camera intrinsic and extrinsic to compute the homography matrix as in Eq.~\ref{eq:homography}. For Epfl-Mpv-VSD, the homography matrix is provided by the original dataset up to a scale, where the scale is inferred using a generic dimension of the objects in the scene, such as door width, and rug size. All distances measured are given in meters.

\subsection{Evaluation Protocol}
\label{sec:eval}
In order to benchmark the proposed VSD method, we only consider the detected people by the pose detector in the groundtruth, disregarding the missed detections by the detector from the entire analysis and thus disentangling the VSD performance from the performance of the detector\footnote{\rev{For the sake of future comparisons, we will release body pose detection results on our dataset as part of the groundtruth.}}

We measure the performance of the algorithm by evaluating its ability to successfully report violations of the SD, namely, considering VSD as a binary classification problem. We report the results of all experiments within the standard metrics used in binary classification evaluation as $Precision = \frac{TP}{TP+FP}$, $Recall = \frac{TP}{TP+FN}$, and $F1-score= 2*\frac{Precision * Recall}{Precision + Recall}$. In this experimental evaluation, we have considered a violation, if the distance between two people in the groundtruth has a value of less than $2m$. In the case, where our VSD algorithm reports that the safe space of a person collides with the safe space of another person in the image, then they are considered as $TP$ cases. Note that $FN$ here means the missed violation of safe distance by the VSD algorithm, rather than by the detector\footnote{In the main paper we only report the F1-score results, and show the corresponding Precision and Recall values in the Supplementary Material.}.

\noindent \textbf{Subjective evaluation on homography estimation:} We have mentioned earlier that our proposed solution is most useful when camera calibration is not available. However, the question might be \textit{``how humans would perform for tuning $\rho_v$ and $\rho_h$, relying purely on their intuitive geometrical understanding of the scene?''}. To answer this question, we formulated a subjective experiment in which we asked 10 human subjects to tune $\rho_v$ and $\rho_h$ parameters till the visual feedback, in the form of projected circles centered at each detected pedestrian onto the image, is reasonable for them. Each subject was asked to tune these parameters once for every single sequence available in each dataset. For the pair of $\rho_v$ and $\rho_h$ selected by each subject for each sequence, we computed the $F1-score$ measure for that sequence and report it in the form of $average \pm std$ among all the subjects, in the Table~\ref{table:quantitative}, `Human estimated' column.

\begin{table*}[ht]
\centering
\small
% \scriptsize
\caption{Investigating the estimation of the Homography matrix $\mathbf{H}$ }
\label{table:quantitative}
\begin{tabular}{c c||c||cc|c||c||c|}
\cline{3-8}
                            &   & \multicolumn{4}{c||}{Proposed VSD} &  \multirow{2}{*}{AutoRect $\mathbf{H}$ \cite{chaudhury2014auto}} &  Monoloco \cite{bertoni2019monoloco} \\
                            \cline{3-6}
                            &    & Human estimated & \multicolumn{3}{c||}{Grid search} & & ( + camera intrinsics)\\
                            \hline\hline
\multicolumn{1}{|l|}{Dataset}                         & Seq.   & F1-score  &  $\rho_h$ & $\rho_v$  & F1-score & F1-score & F1-score \\ \hline \hline 
\multicolumn{1}{|l|}{\multirow{4}{*}{EPFL-MPV}}       & C0     & 77.43 $\pm$ 1.12 &  0.7  & 0.5 & \textbf{77.90} & 73.47 & 73.46 (N.A.) \\
\multicolumn{1}{|l|}{}                                & C1     & 71.24 $\pm$ 4.92 &  0.5  & 0.6 & \textbf{75.39} & 61.17 & 70.19 (N.A.) \\ 
\multicolumn{1}{|l|}{}                                & C2     & 75.69 $\pm$ 3.87 &  0.6  & 0.6 & \textbf{78.67} & 78.14  & 77.20 (N.A.) \\ 
\multicolumn{1}{|l|}{}                                & C3     & 73.34 $\pm$ 3.51 &  0.5  & 0.6 & \textbf{75.86} & 58.36 & 72.60 (N.A.) \\ \cline{2-8}
\multicolumn{1}{|l|}{}                                & Avg.   & 74.42 $\pm$ 3.35 &   -   &  -  & \textbf{76.94} & 67.75 & 73.36 (N.A.) \\ \hline \hline
\multicolumn{1}{|l|}{\multirow{7}{*}{EPFL-wildtrack}} & C1     & 84.78 $\pm$ 1.11 &  0.8  & 0.7 & \textbf{86.31} & 61.80 & 70.07 (70.50)  \\ 
\multicolumn{1}{|l|}{}                                & C2     & 83.84 $\pm$ 1.19 &  0.8  & 0.6 & \textbf{85.57} & 57.27 & 68.31 (68.11)  \\ 
\multicolumn{1}{|l|}{}                                & C3     & 85.52 $\pm$ 2.08 &  0.8  & 0.7 & \textbf{87.96} & 45.21 & 66.73 (67.10)  \\ 
\multicolumn{1}{|l|}{}                                & C4     & 81.10 $\pm$ 4.03 &  0.6  & 0.8 & \textbf{85.54} & 35.06 & 64.29 (65.13)  \\  
\multicolumn{1}{|l|}{}                                & C5     & 67.63 $\pm$ 3.63 &  0.8  & 0.8 & \textbf{69.91} & 50.99 & 54.80 (54.69)  \\  
\multicolumn{1}{|l|}{}                                & C6     & 62.87 $\pm$ 2.23 &  0.8  & 0.8 & \textbf{65.27} & 39.54 & 49.64 (49.58)  \\ 
\multicolumn{1}{|l|}{}                                & C7     & 84.46 $\pm$ 3.51 &  0.5  & 0.7 & \textbf{86.96} & 55.63 & 68.44 (68.74)  \\ \cline{2-8}
\multicolumn{1}{|l|}{}                                & Avg.   & 78.60 $\pm$ 2.54 &  -    &  -  & \textbf{81.07} & 49.77 & 63.18 (63.40)   \\ \hline \hline
\multicolumn{1}{|l|}{OxTown}                          &  -     & 76.94 $\pm$ 4.52 &  0.5  & 0.8 & \textbf{81.04} & 51.78 & 54.57 (54.54)  \\ \hline
\end{tabular}
\end{table*}

\begin{table*}[ht]
\centering
\small
% \scriptsize
\caption{Investigating metric reference choice, using the fixed homography-related parameters (the best choice among subjects).}
\label{table:quantitative2}
\begin{tabular}{c c||cc||c|c|c||c|}
\cline{3-8}
                            &   &  & &  \multicolumn{3}{c||}{Body parts} & \multirow{2}{*}{BBX height} \\
                            \cline{5-7}
                            &    &  \multicolumn{2}{c||}{ \multirow{2}{*}{ Ratios}}  & Leg length  & Arm length &  Torso length &  \\ 
                            &    & & & (0.85  m)  & (0.70 m) &  (0.60 m) & (1.7 m) \\ \hline\hline
\multicolumn{1}{|l|}{Dataset}                         & Seq.   &  $\rho_h$ & $\rho_v$  & F1-score  & F1-score & F1-score & F1-score\\ \hline \hline 
\multicolumn{1}{|l|}{\multirow{4}{*}{PFL-MPV}}       & C0      &  0.6  & 0.5 & 77.14 & 74.68 & \textbf{77.64} & 76.71 \\
\multicolumn{1}{|l|}{}                                & C1     &  0.5  & 0.6 & 74.80 & 71.19 & \textbf{75.38} & 73.90 \\ 
\multicolumn{1}{|l|}{}                                & C2     & 0.8   & 0.5 & 78.43 & 70.02 & 77.12 & \textbf{78.63} \\ 
\multicolumn{1}{|l|}{}                                & C3     & 0.8   & 0.5 & 73.41 & 66.38 & 71.19 & \textbf{74.69} \\ \cline{2-8}
\multicolumn{1}{|l|}{}                                & Avg.   &  - &  -  & 75.94 & 70.56 & \textbf{75.98} & 75.96 \\ \hline \hline
\multicolumn{1}{|l|}{\multirow{7}{*}{EPFL-wildtrack}} & C1     &  0.5  & 0.8 & 83.93 & 81.46 & \textbf{84.10} & 79.18 \\ 
\multicolumn{1}{|l|}{}                                & C2     &  0.7  & 0.5 & 81.07 & 80.47 & \textbf{82.64} & 81.18 \\ 
\multicolumn{1}{|l|}{}                                & C3     &  0.5  & 0.8 & 83.57 & 81.00 & \textbf{85.37} & 79.13 \\ 
\multicolumn{1}{|l|}{}                                & C4     &  0.6  & 0.8 & 85.95 & 73.30 & \textbf{86.68} & 81.52 \\  
\multicolumn{1}{|l|}{}                                & C5     &  0.8  & 0.8 & 68.40 & \textbf{72.58} & 69.82 & 64.65 \\  
\multicolumn{1}{|l|}{}                                & C6     &  0.8  & 0.7 & 62.98 & 63.53 & \textbf{63.72} & 59.73 \\ 
\multicolumn{1}{|l|}{}                                & C7     &  0.6  & 0.8 & \textbf{87.20} & 76.95 & 84.03 & 86.62 \\ \cline{2-8}
\multicolumn{1}{|l|}{}                                & Avg.   &  -    &  -  & 79.01 & 75.61 & \textbf{79.48} & 76.00 \\ \hline \hline
\multicolumn{1}{|l|}{OxTown}                          &  -     &  0.5  & 0.8 & \textbf{82.59} & 73.03 & 81.04 & 72.38 \\ \hline
\end{tabular}
\end{table*}

\noindent \textbf{Homography parameters tuning with grid search:} This experiment aims to answer \textit{``what is the best achievable performance using our proposed algorithm?''}. To this end, we tune the optimal $\rho_v$ and $\rho_h$ values by grid searching separately over each sequence. 
We discretise the search space of $\rho_v$ and $\rho_h$ $\in [0,1]$ with a step $0.1$. For each sequence, the combination of $\rho_v$ and $\rho_h$ that produces the best performance is then selected as the optimal parameters. In Table~\ref{table:quantitative}, `Grid search' column, we report the quantitative results per each sequence using the optimal parameters. 

\noindent \textbf{Comparison with automatic homography estimation:} \rev{Last, we benchmark the proposed VSD pipeline against two methods. AutoRect \cite{chaudhury2014auto} provides a model for single image automatic homography estimation. Monoloco \cite{bertoni2019monoloco} is a method for monocular 3D human localization from a single image. To compare our proposed model with AutoRect, the circular safe space around each detected subject on the scene is projected onto the ground plane through an estimate of $\mathbf{H}$, computed by evaluating vanishing points and projecting them to infinity, while the $\mathbf{H}$ is estimated real time for each frame of the sequence. The results of this experiment are listed in Table~\ref{table:quantitative}, column `AutoRect $\mathbf{H}$'. In the last column of Table~\ref{table:quantitative}, the results of applying Monoloco on each sequence of the datasets are reported. We show results using the `default' camera intrinsic matrix $\mathbf{K}$ provided by the Monoloco algorithm, and the groundtruth $\mathbf{K}$ when available (Epfl-Wildtrack-VSD and  OxTown-VSD). Note that all the experiments in Table \ref{table:quantitative}, except Monoloco, the metric reference is set to the `Torso' for coherence.} 

\noindent \textbf{Ablation study on metric reference:} As discussed in Sec.~\ref{sec:pose-height}, pose detectors facilitate a better estimation of metric reference with regards to human body detection as a \rev{bounding box}. In Table~\ref{table:quantitative2}, `BBX height' column we report the results of our proposed algorithm using the height of the \rev{bounding box} (obtained using OpenPose joints as detailed in Sec.~\ref{sec:person_localisation}). As inferring the length of any of the body parts from a single \rev{bounding box} is implausible, we consider the height of the detected \rev{bounding box} as the average height of a person ($1.70 m$) and from there we estimate the radius $r$ of the safe space. Moreover, we also validate  metric references using various body parts, namely, `Leg', `Arm', and `Torso'. More details on the actual OpenPose joints used to infer such pixel lengths can be found in the Supplementary Material. The obtained results from this experiment can be found in Table~\ref{table:quantitative2} in the respective columns. Each body part is associated with a metric reference according to population statistics \cite{murtinho2015leonardo,human_ratio_nasa}, as detailed in the table. For the sake of fair comparisons, $\rho_v$ and $\rho_h$ are kept the same for this ablation study according to the best subjective estimation.

\begin{figure}[h!]
\begin{center}
	\begin{tabular}{@{}c@{}c@{}c}
		\includegraphics[height=0.17\textwidth,keepaspectratio]{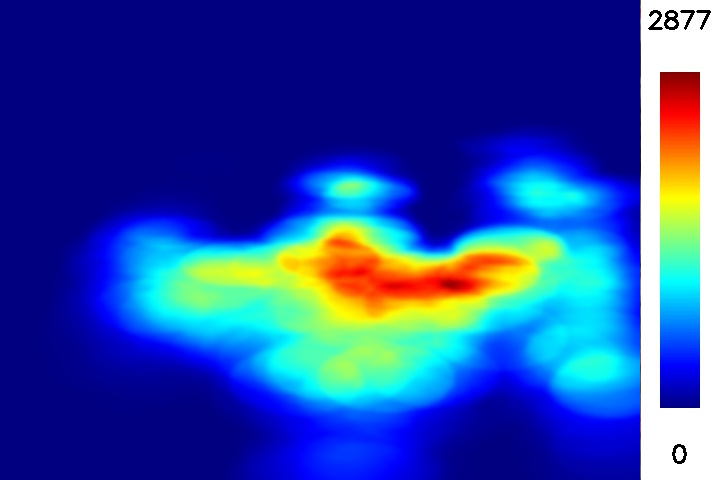}&\includegraphics[height=0.17\textwidth,keepaspectratio]{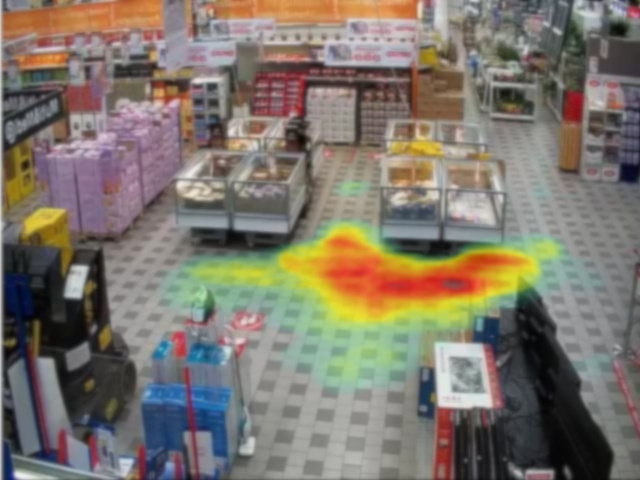}\\
		(a)  & (b) 
   	\end{tabular}
\end{center}
\caption{The heatmap in the images represents the areas on which SD violations mostly occur. (a) the maximum number of observed violations is depicted in red and overlaid in the image frame in (b). The heatmap is produced using observations collected in 1-hour from a monitored retail shop. The area with the most observed violations is due to an exhibit placed in the main entrance of the shop that causes increasing SD violations.}
\label{fig:retail}
\end{figure}

\subsection{Experimental Results Discussion}
\label{sec:discussion}
As it can be extrapolated from Table~\ref{table:quantitative}, the best average obtained results for all datasets is achieved by grid-searching the best parameters. Our original proposal to rely on human geometrical understanding to estimate $\rho_v$ and $\rho_h$ (Table~\ref{table:quantitative},`Human estimated' column) performs only slightly lower than the best choice of parameters (Table~\ref{table:quantitative}, `Grid search' column). Moreover, the relatively low $std$ of the results in most of the sequences is an indication of the inter-raters agreement on the \textit{correctness} measures. Besides, the proposed VSD algorithm, either requiring manual tuning of parameters or validating them, on average per dataset, performs better than applying automatic estimation of $\mathbf{H}$ using AutoRect. \rev{Monoloco provides instead the 3D localization starting from pose detections and the matrix for the intrinsic camera parameters $\mathbf{K}$. If $\mathbf{K}$ Monoloco has a default value. It can be seen that using such a default $\mathbf{K}$ or the  groundtruth one does not lead to relevant differences, indicating that Monoloco is not very sensitive to it in the evaluated datasets. Instead, the Monoloco network is probably very biased towards its training data since our proposed algorithm outperforms it in all the sequences (although it outperforms AutoRect).}

\rev{Concerning Table \ref{table:quantitative2}, it can be seen that on average using \rev{bounding boxes} as the metric reference is not a reliable option. This observation emphasizes the importance of having information regarding human body joints instead of a mere \rev{bounding box}, as already discussed. The last observation regards the choice of different body parts, namely 'Leg length', `Arm length', or 'Torso length' for inferring distances in the image. We observe that torso provides a more robust reference, in general performing better than arm and leg. This is reasonable since, in normal conditions in which people are walking in the scene, legs and arms can undergo various deformations due to the multiple numbers of joints in them, while the torso is merely one straight line which connects the neck to the heaps. For OxTown only, the leg reference performs slightly better than the torso one. Arm is not a good reference body part as it often undergoes occlusions with torso, preventing from a correct measurement of the metric reference. Additional qualitative results in the form of videos are provided  in the supplementary material.}

\subsection{Real-world Scenario Challenges and Next Steps}
\label{sec:real-scenario}

Our proposed setup is currently able to monitor up to 18 cameras with 640x480 resolution, running on a single Alienware Aurora R8 PC equipped with an NVidia RTX2080 GPU at 1 fps. The system has been successfully applied for monitoring social distancing 24/7 in challenging scenarios such as corporate offices, airports, and shopping malls, where more than 100,000 observations per hour are being evaluated. The collected statistics are used for both generating real time alarms finalised to break apart big groups of persons and for generating statistics in order to achieve effective space redesigns (see Fig. \ref{fig:retail}).

The collection of this huge stream of data on real application scenarios allows us to highlight the most important open issues that future developments should focus on:
\begin{itemize}[noitemsep,topsep=0pt]
  \item System alarms for the violations of safe spaces should not be based only on single frame analysis, but rather encompass temporal information and people tracking.
  \item Currently many of the SD laws allow family members to stay close to each other. A robust system ideally should be able to recognise family member formations and discard their violations from the computed ones.
 \item Our approach is frame-based and it can be applied to any kind of camera under certain setting constraints. In the case of fixed cameras, a statistical approach can be used for inferring metric references. This statistical approach would allow the system to be more robust to outliers (such as kids on the scene).
\end{itemize}

\section{Conclusion}

We provided the first$^2$ VSD method for estimating SD from single uncalibrated images together with a \rev{augmented VSD annotations for three public} datasets for further VSD evaluation purposes. The proposed method performs favorably against competing baselines, showing the importance of using a pose estimator instead of a bounding box detector. This allows exploiting human body parts as robust metric references; in particular, we find the length of the torso as the most reliable. In addition, we demonstrated the competence of the proposed semi-automatic homography estimation against automatic estimation using state-of-the-art. \rev{Our proposed method might fail when the underlying pose detector fails to properly detect the body joints due to its strong dependence on the joint positional information. The promising performance of the proposed method is also guaranteed under certain camera positioning which limits its applicability to surveillance settings in particular. In the same line of reasoning, our semi-automatic method is only capable of estimating one homography per scene which leads to deteriorations in its performance when multiple homographies are present in the scene.}
Future work will provide a more in-depth analysis of the causes of SD violations (e.g. presence of families, use of personal protective equipment) and the development of forecasting methods to avoid the formation of gatherings. 

%Moreover, as mentioned beforehand, further research efforts should propose  robust metric reference from people body parts. 
% In particular, we plan working on accumulating image-projected persons' heights (in pixels) in order to be able to calculate person median heights for as much image coordinates as possible and use Ransac to fit a plane on them. Having fit the median heights plane, we can use it for calculating the safety distance of any new detected persons, using his/her feet position only.

{\small
\bibliographystyle{ieee_fullname}
\bibliography{2021_wacv_vsd}

\begin{thebibliography}{10}\itemsep=-1pt

\bibitem{Abbas:az:2019}
S.~A. {Abbas} and A. {Zisserman}.
\newblock A geometric approach to obtain a bird's eye view from an image.
\newblock In {\em 2019 IEEE/CVF International Conference on Computer Vision
  Workshop (ICCVW)}, pages 4095--4104, 2019.

\bibitem{human_ratio_nasa}
National Aeronautics and Space Administration.
\newblock Man systems integration standards.
\newblock Technical report, 1987.

\bibitem{baten2014human}
Joerg Baten and Matthias Blum.
\newblock Human height since 1820.
\newblock 2014.

\bibitem{bazin20123}
Jean-Charles Bazin and Marc Pollefeys.
\newblock 3-line ransac for orthogonal vanishing point detection.
\newblock In {\em 2012 IEEE/RSJ International Conference on Intelligent Robots
  and Systems}, pages 4282--4287. IEEE, 2012.

\bibitem{bazin2012globally}
Jean-Charles Bazin, Yongduek Seo, C{\'e}dric Demonceaux, Pascal Vasseur,
  Katsushi Ikeuchi, Inso Kweon, and Marc Pollefeys.
\newblock Globally optimal line clustering and vanishing point estimation in
  manhattan world.
\newblock In {\em 2012 IEEE Conference on Computer Vision and Pattern
  Recognition}, pages 638--645. IEEE, 2012.

\bibitem{benabdelkader2008statistical}
Chiraz BenAbdelkader and Yaser Yacoob.
\newblock Statistical body height estimation from a single image.
\newblock In {\em 2008 8th IEEE International Conference on Automatic Face \&
  Gesture Recognition}, pages 1--7. IEEE, 2008.

\bibitem{benenson2014ten}
Rodrigo Benenson, Mohamed Omran, Jan Hosang, and Bernt Schiele.
\newblock Ten years of pedestrian detection, what have we learned?
\newblock In {\em European Conference on Computer Vision}, pages 613--627.
  Springer, 2014.

\bibitem{OxTown2011}
Ben Benfold and Ian Reid.
\newblock Stable multi-target tracking in real-time surveillance video.
\newblock In {\em CVPR}, pages 3457--3464, June 2011.

\bibitem{bertoni2019monoloco}
Lorenzo Bertoni, Sven Kreiss, and Alexandre Alahi.
\newblock Monoloco: Monocular 3d pedestrian localization and uncertainty
  estimation.
\newblock In {\em Proceedings of the IEEE International Conference on Computer
  Vision}, pages 6861--6871, 2019.

\bibitem{bogo2016keep}
Federica Bogo, Angjoo Kanazawa, Christoph Lassner, Peter Gehler, Javier Romero,
  and Michael~J Black.
\newblock Keep it smpl: Automatic estimation of 3d human pose and shape from a
  single image.
\newblock In {\em European Conference on Computer Vision}, pages 561--578.
  Springer, 2016.

\bibitem{boominathan2016crowdnet}
Lokesh Boominathan, Srinivas~SS Kruthiventi, and R~Venkatesh Babu.
\newblock Crowdnet: A deep convolutional network for dense crowd counting.
\newblock In {\em ACM International Conference on Multimedia}, pages 640--644,
  2016.

\bibitem{brooks2020psychological}
Samantha~K Brooks, Rebecca~K Webster, Louise~E Smith, Lisa Woodland, Simon
  Wessely, Neil Greenberg, and Gideon~James Rubin.
\newblock The psychological impact of quarantine and how to reduce it: rapid
  review of the evidence.
\newblock {\em The Lancet}, 2020.

\bibitem{openpose:ijcv}
Z. {Cao}, G. {Hidalgo Martinez}, T. {Simon}, S. {Wei}, and Y.~A. {Sheikh}.
\newblock Openpose: Realtime multi-person 2d pose estimation using part
  affinity fields.
\newblock {\em IEEE Transactions on Pattern Analysis and Machine Intelligence},
  2019.

\bibitem{chaudhury2014auto}
Krishnendu Chaudhury, Stephen DiVerdi, and Sergey Ioffe.
\newblock Auto-rectification of user photos.
\newblock In {\em 2014 IEEE International Conference on Image Processing
  (ICIP)}, pages 3479--3483. IEEE, 2014.

\bibitem{wildtrack2018}
Tatjana Chavdarova, Pierre Baqué, Andrii Maksai, Stéphane Bouquet, Cijo Jose,
  Louis Lettry, Francois Fleuret, Pascal Fua, and Luc~Van Gool.
\newblock Wildtrack: A multi-camera hd dataset for dense unscripted pedestrian
  detection.
\newblock In {\em Proc. of IEEE/CVF Conference On Computer Vision And Pattern
  Recognition}, pages 5030--5039, New York, 2018. IEEE.

\bibitem{criminisi2000single}
Antonio Criminisi, Ian Reid, and Andrew Zisserman.
\newblock Single view metrology.
\newblock {\em International Journal of Computer Vision}, 40(2):123--148, 2000.

\bibitem{vsd2020}
M. {Cristani}, A.~D. {Bue}, V. {Murino}, F. {Setti}, and A. {Vinciarelli}.
\newblock The visual social distancing problem.
\newblock {\em IEEE Access}, 8:126876--126886, 2020.

\bibitem{dey2014estimating}
Ratan Dey, Madhurya Nangia, Keith~W Ross, and Yong Liu.
\newblock Estimating heights from photo collections: A data-driven approach.
\newblock In {\em Proceedings of the second ACM conference on Online social
  networks}, pages 227--238, 2014.

\bibitem{Epfl-mpv2008}
Francois Fleuret, Jerome Berclaz, Richard Lengagne, and Pascal Fua.
\newblock Multicamera people tracking with a probabilistic occupancy map.
\newblock {\em IEEE Trans. Pattern Anal. Mach. Intell.}, 30(2):267–282, Feb.
  2008.

\bibitem{ge2020ps}
Zheng Ge, Zequn Jie, Xin Huang, Rong Xu, and Osamu Yoshie.
\newblock Ps-rcnn: Detecting secondary human instances in a crowd via primary
  object suppression.
\newblock {\em arXiv preprint arXiv:2003.07080}, 2020.

\bibitem{golda2019human}
Thomas Golda, Tobias Kalb, Arne Schumann, and J{\"u}rgen Beyerer.
\newblock Human pose estimation for real-world crowded scenarios.
\newblock In {\em 2019 16th IEEE International Conference on Advanced Video and
  Signal Based Surveillance (AVSS)}, pages 1--8. IEEE, 2019.

\bibitem{guler2018densepose}
R{\i}za~Alp G{\"u}ler, Natalia Neverova, and Iasonas Kokkinos.
\newblock Densepose: Dense human pose estimation in the wild.
\newblock In {\em Proceedings of the IEEE Conference on Computer Vision and
  Pattern Recognition}, pages 7297--7306, 2018.

\bibitem{Gunel_2019_ICCV}
Semih Gunel, Helge Rhodin, and Pascal Fua.
\newblock What face and body shapes can tell us about height.
\newblock In {\em The IEEE International Conference on Computer Vision (ICCV)
  Workshops}, Oct 2019.

\bibitem{Hartley2004}
R.~I. Hartley and A. Zisserman.
\newblock {\em Multiple View Geometry in Computer Vision}.
\newblock Cambridge University Press, ISBN: 0521540518, second edition, 2004.

\bibitem{hoiem2007recovering}
Derek Hoiem, Alexei~A Efros, and Martial Hebert.
\newblock Recovering surface layout from an image.
\newblock {\em International Journal of Computer Vision}, 75(1):151--172, 2007.

\bibitem{hold2018perceptual}
Yannick Hold-Geoffroy, Kalyan Sunkavalli, Jonathan Eisenmann, Matthew Fisher,
  Emiliano Gambaretto, Sunil Hadap, and Jean-Fran{\c{c}}ois Lalonde.
\newblock A perceptual measure for deep single image camera calibration.
\newblock In {\em Proceedings of the IEEE Conference on Computer Vision and
  Pattern Recognition}, pages 2354--2363, 2018.

\bibitem{johnson2020social}
Joseph Johnson~Jr, Shiblee Hasan, David Lee, Chris Hluchan, and Nazia Ahmed.
\newblock Social-distancing monitoring using portable electronic devices.
\newblock 2020.

\bibitem{lee2000monitoring}
Lily Lee, Raquel Romano, and Gideon Stein.
\newblock Monitoring activities from multiple video streams: Establishing a
  common coordinate frame.
\newblock {\em IEEE Transactions on pattern analysis and machine intelligence},
  22(8):758--767, 2000.

\bibitem{li2019crowdpose}
Jiefeng Li, Can Wang, Hao Zhu, Yihuan Mao, Hao-Shu Fang, and Cewu Lu.
\newblock Crowdpose: Efficient crowded scenes pose estimation and a new
  benchmark.
\newblock In {\em Proceedings of the IEEE Conference on Computer Vision and
  Pattern Recognition}, pages 10863--10872, 2019.

\bibitem{liu2011surveillance}
Jingchen Liu, Robert~T Collins, and Yanxi Liu.
\newblock Surveillance camera autocalibration based on pedestrian height
  distributions.
\newblock In {\em Proceedings of the British Machine Vision Conference}, page
  144, 2011.

\bibitem{liu2018decidenet}
Jiang Liu, Chenqiang Gao, Deyu Meng, and Alexander~G Hauptmann.
\newblock Decidenet: Counting varying density crowds through attention guided
  detection and density estimation.
\newblock In {\em IEEE Conference on Computer Vision and Pattern Recognition
  (CVPR)}, pages 5197--5206, 2018.

\bibitem{Liu_2019_CVPR}
Songtao Liu, Di Huang, and Yunhong Wang.
\newblock Adaptive nms: Refining pedestrian detection in a crowd.
\newblock In {\em The IEEE Conference on Computer Vision and Pattern
  Recognition (CVPR)}, June 2019.

\bibitem{Lopez_CVPR_2019}
Manuel L\'opez-Antequera, Roger Mar\'i, Pau Gargallo, Yubin Kuang, Javier
  Gonzalez-Jimenez, and Gloria Haro.
\newblock Deep single image camera calibration with radial distortion.
\newblock In {\em Computer Vision and Pattern Recognition (CVPR)}, June 2019.

\bibitem{lu20172}
Xiaohu Lu, Jian Yaoy, Haoang Li, and Yahui Liu.
\newblock 2-line exhaustive searching for real-time vanishing point estimation
  in manhattan world.
\newblock In {\em 2017 IEEE Winter Conference on Applications of Computer
  Vision (WACV)}, pages 345--353. IEEE, 2017.

\bibitem{lv2006camera}
Fengjun Lv, Tao Zhao, and Ramakant Nevatia.
\newblock Camera calibration from video of a walking human.
\newblock {\em IEEE transactions on pattern analysis and machine intelligence},
  28(9):1513--1518, 2006.

\bibitem{groundnet:2019}
Yunze Man, Xinshuo Weng, Xi Li, and Kris Kitani.
\newblock Groundnet: Monocular ground plane normal estimation with geometric
  consistency.
\newblock In {\em Proceedings of the 27th ACM International Conference on
  Multimedia}, page 2170–2178, 2019.

\bibitem{martinez2017simple}
Julieta Martinez, Rayat Hossain, Javier Romero, and James~J Little.
\newblock A simple yet effective baseline for 3d human pose estimation.
\newblock In {\em Proceedings of the IEEE International Conference on Computer
  Vision}, pages 2640--2649, 2017.

\bibitem{mehta2019xnect}
Dushyant Mehta, Oleksandr Sotnychenko, Franziska Mueller, Weipeng Xu, Mohamed
  Elgharib, Pascal Fua, Hans-Peter Seidel, Helge Rhodin, Gerard Pons-Moll, and
  Christian Theobalt.
\newblock Xnect: Real-time multi-person 3d human pose estimation with a single
  rgb camera.
\newblock {\em arXiv preprint arXiv:1907.00837}, 2019.

\bibitem{mehta2017vnect}
Dushyant Mehta, Srinath Sridhar, Oleksandr Sotnychenko, Helge Rhodin, Mohammad
  Shafiei, Hans-Peter Seidel, Weipeng Xu, Dan Casas, and Christian Theobalt.
\newblock Vnect: Real-time 3d human pose estimation with a single rgb camera.
\newblock {\em ACM Transactions on Graphics (TOG)}, 36(4):1--14, 2017.

\bibitem{mirzaei2011optimal}
Faraz~M Mirzaei and Stergios~I Roumeliotis.
\newblock Optimal estimation of vanishing points in a manhattan world.
\newblock In {\em 2011 International Conference on Computer Vision}, pages
  2454--2461. IEEE, 2011.

\bibitem{moreno20173d}
Francesc Moreno-Noguer.
\newblock 3d human pose estimation from a single image via distance matrix
  regression.
\newblock In {\em Proceedings of the IEEE Conference on Computer Vision and
  Pattern Recognition}, pages 2823--2832, 2017.

\bibitem{murtinho2015leonardo}
Vitor Murtinho.
\newblock Leonardo’s vitruvian man drawing: a new interpretation looking at
  leonardo’s geometric constructions.
\newblock {\em Nexus Network Journal}, 17(2):507--524, 2015.

\bibitem{rangel2017entropy}
A Rangel-Huerta, AL Ballinas-Hern{\'a}ndez, and A Mu{\~n}oz-Mel{\'e}ndez.
\newblock An entropy model to measure heterogeneity of pedestrian crowds using
  self-propelled agents.
\newblock {\em Physica A: Statistical Mechanics and its Applications},
  473:213--224, 2017.

\bibitem{rogez2019lcr}
Gr{\'e}gory Rogez, Philippe Weinzaepfel, and Cordelia Schmid.
\newblock Lcr-net++: Multi-person 2d and 3d pose detection in natural images.
\newblock {\em IEEE transactions on pattern analysis and machine intelligence},
  2019.

\bibitem{rother2002new}
Carsten Rother.
\newblock A new approach to vanishing point detection in architectural
  environments.
\newblock {\em Image and Vision Computing}, 20(9-10):647--655, 2002.

\bibitem{setti2018count}
Francesco Setti, Davide Conigliaro, Michele Tobanelli, and Marco Cristani.
\newblock Count on me: learning to count on a single image.
\newblock {\em IEEE Transactions on Circuits and Systems for Video Technology},
  28(8):1798--1806, 2018.

\bibitem{sindagi2017cnn}
Vishwanath~A Sindagi and Vishal~M Patel.
\newblock Cnn-based cascaded multi-task learning of high-level prior and
  density estimation for crowd counting.
\newblock In {\em IEEE International Conference on Advanced Video and Signal
  Based Surveillance (AVSS)}, pages 1--6. IEEE, 2017.

\bibitem{sindagi2018survey}
Vishwanath~A Sindagi and Vishal~M Patel.
\newblock A survey of recent advances in cnn-based single image crowd counting
  and density estimation.
\newblock {\em Pattern Recognition Letters}, 107:3--16, 2018.

\bibitem{szeliski2011}
R. Szeliski.
\newblock {\em Computer Vision: Algorithms and Applications}.
\newblock Springer International Publishing, 2011.

\bibitem{tang2019esther}
Zheng Tang, Yen-Shuo Lin, Kuan-Hui Lee, Jenq-Neng Hwang, and Jen-Hui Chuang.
\newblock Esther: Joint camera self-calibration and automatic radial distortion
  correction from tracking of walking humans.
\newblock {\em IEEE Access}, 7:10754--10766, 2019.

\bibitem{padnet2020}
Y. {Tian}, Y. {Lei}, J. {Zhang}, and J.~Z. {Wang}.
\newblock Padnet: Pan-density crowd counting.
\newblock {\em IEEE Transactions on Image Processing}, 29:2714--2727, 2020.

\bibitem{tome2017lifting}
Denis Tome, Chris Russell, and Lourdes Agapito.
\newblock Lifting from the deep: Convolutional 3d pose estimation from a single
  image.
\newblock In {\em Proceedings of the IEEE Conference on Computer Vision and
  Pattern Recognition}, pages 2500--2509, 2017.

\bibitem{vandoni2019evidential}
Jennifer Vandoni, Emanuel Aldea, and Sylvie Le~H{\'e}garat-Mascle.
\newblock Evidential query-by-committee active learning for pedestrian
  detection in high-density crowds.
\newblock {\em International Journal of Approximate Reasoning}, 104:166--184,
  2019.

\bibitem{vester2012estimating}
Johan Vester.
\newblock Estimating the height of an unknown object in a 2d image.
\newblock {\em KTH CSIC}, 2012.

\bibitem{wagner2019}
Robert Wagner, Daniel Crispell, Patrick Feeney, and Joe Mundy.
\newblock 4-d scene alignment in surveillance video.
\newblock {\em arXiv preprint arXiv:1906.01675}, 2019.

\bibitem{wang2014robust}
Chunyu Wang, Yizhou Wang, Zhouchen Lin, Alan~L Yuille, and Wen Gao.
\newblock Robust estimation of 3d human poses from a single image.
\newblock In {\em Proceedings of the IEEE Conference on Computer Vision and
  Pattern Recognition}, pages 2361--2368, 2014.

\bibitem{Wang_2018_CVPR}
Xinlong Wang, Tete Xiao, Yuning Jiang, Shuai Shao, Jian Sun, and Chunhua Shen.
\newblock Repulsion loss: Detecting pedestrians in a crowd.
\newblock In {\em The IEEE Conference on Computer Vision and Pattern
  Recognition (CVPR)}, June 2018.

\bibitem{watson-2020-footprints}
Jamie Watson, Michael Firman, Aron Monszpart, and Gabriel~J. Brostow.
\newblock Footprints and free space from a single color image.
\newblock In {\em Computer Vision and Pattern Recognition ({CVPR})}, 2020.

\bibitem{wildenauer2012robust}
Horst Wildenauer and Allan Hanbury.
\newblock Robust camera self-calibration from monocular images of manhattan
  worlds.
\newblock In {\em 2012 IEEE Conference on Computer Vision and Pattern
  Recognition}, pages 2831--2838. IEEE, 2012.

\bibitem{detection_depth}
L. {Xia}, C. {Chen}, and J.~K. {Aggarwal}.
\newblock Human detection using depth information by kinect.
\newblock In {\em Proc. of Computer Vision and Pattern Recognition Workshops},
  pages 15--22, 2011.

\bibitem{xu2020estimating}
Yan Xu, Vivek Roy, and Kris Kitani.
\newblock Estimating 3d camera pose from 2d pedestrian trajectories.
\newblock In {\em 2020 IEEE Winter Conference on Applications of Computer
  Vision (WACV)}, pages 2568--2577. IEEE, 2020.

\bibitem{pseudo_lidar++}
Yurong You, Yan Wang, Wei-Lun Chao, Divyansh Garg, Geoff Pleiss, Bharath
  Hariharan, Mark Campbell, and Kilian~Q Weinberger.
\newblock Pseudo-lidar++: Accurate depth for 3d object detection in autonomous
  driving.
\newblock In {\em Proc. of International Conference on Learning
  Representations}, 2020.

\bibitem{zhang2016vanishing}
Lilian Zhang, Huimin Lu, Xiaoping Hu, and Reinhard Koch.
\newblock Vanishing point estimation and line classification in a manhattan
  world with a unifying camera model.
\newblock {\em International Journal of Computer Vision}, 117(2):111--130,
  2016.

\bibitem{zhou2017towards}
Xingyi Zhou, Qixing Huang, Xiao Sun, Xiangyang Xue, and Yichen Wei.
\newblock Towards 3d human pose estimation in the wild: a weakly-supervised
  approach.
\newblock In {\em Proceedings of the IEEE International Conference on Computer
  Vision}, pages 398--407, 2017.

\end{thebibliography}
}

\appendix{}

\section{Supplementary Material}

We provide in this Supplementary Material more information about the experimental evaluation in the paper, in particular, related to Precision and Recall for the three datasets, \textit{Epfl-Mpv-VSD}, \textit{Epfl-Wildtrack-VSD}, and  \textit{OxTown-VSD}. In addition, we include the visualisation of heathmaps that shows intuitively the image areas where the violations appear more. 

Finally, we provide short video clips for all the sequences in our dataset with the qualitative results of the proposed algorithm overlaid. It is worth to mention that the videos are generated given the best $\rho_h$ and $\rho_v$ by grid-search strategy (given in Table 1 of the main manuscript) and the best performing body part for each sequence (given in Table 2 of the main manuscript).\footnote{We do not visualise the whole dataset images but only a fraction due to the overall size.}     

\section{Relation between homography matrix to the 2 trapezoid ratios}
\begin{figure}
	\centering
	\includegraphics[width=\columnwidth]{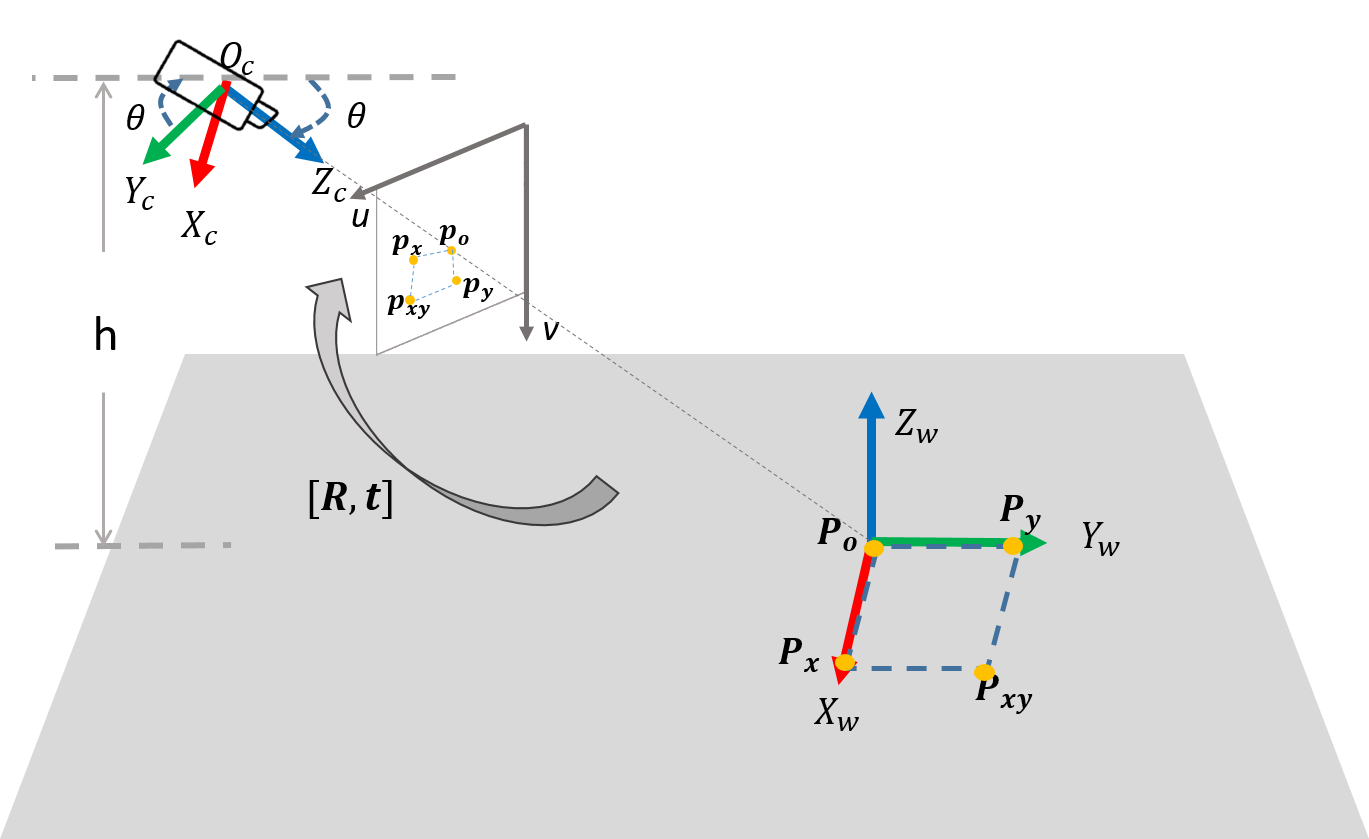}
	\caption{Illustration of the world and camera coordinate system under our assumptions that i) the camera only has a tilt angle $\theta$ with zero roll and pan angle, and ii) the camera has the height $h$ from the ground and the world origin locates at where the camera principal axis intersects on the ground plane. The $\mathbf{P}$ on the ground plane can be mapped to the pixel $\mathbf{p}$ on the image through the homography matrix. On the ground plane the corner points on the unit square $\mathbf{P}_{o}$, $\mathbf{P}_{x}$, $\mathbf{P}_{y}$ and $\mathbf{P}_{xy}$ correspond to the pixel points $\mathbf{p}_{o}$, $\mathbf{p}_{x}$, $\mathbf{p}_{y}$ and $\mathbf{p}_{xy}$, respectively.}
	\label{fig:camera_setup_ratio}
\end{figure}
Let us consider a pinhole camera model with the camera intrinsic matrix $\mathbf{K}\in \mathbb{R}^{3\times3}$
and extrinsic matrix $[\mathbf{R}|\mathbf{t}]$. 
Any 2D point in world ground plane $\mathbf{P}=\left[X, Y, 1\right]^T$ can then be projected to the image plane at pixel position $\mathbf{p}= \left[u, v, 1\right]^T$ as:
\begin{equation}
\centering
s\bold{p} = \bold{K}\begin{bmatrix}
\bold{r}_{1} & \bold{r}_{2} & \bold{t}
\end{bmatrix}\bold{P} = \bold{H}\bold{P},
\label{eq:homography}
\end{equation}
where $s$ is the scale factor, $\bold{H}$ is the {\em homography matrix} to project points on the ground plane to the image plane and $\bold{P}' = \left[X, Y, 1\right]^T$.

Based on our assumption of the world coordinate with the camera roll angle and pan angle to be $0^{\circ}$ and the camera origin, we will have:
\begin{equation}
\centering
\textbf{H} = \textbf{K}\begin{bmatrix}
\textbf{r}_{1} & \textbf{r}_{2} & \textbf{t}
\end{bmatrix} = \textbf{K} \begin{bmatrix}
1 & 0 & o\\ 
0 & -\cos(\theta) & -\frac{h}{tan(\theta)}\\ 
0 & -\sin(\theta) & h
\end{bmatrix}
\label{eq:homography_parameter}
\end{equation}

As we mentioned in the main paper, lines that are parallel to the X-axis in the ground plane remain horizontal in the projected image plane, and lines parallel to the Y-axis converge to the vanishing point along Y-axis in the image plane. Therefore, a rectangular area on the ground plane with a width $W$ and a height $H$ would be projected into an isosceles trapezoidal shape with short-based width $W^{'}$ and height $H^{'}$, where the horizontal ratio $\rho_{h} = \frac{W^{'}}{W}$ and vertical ratio $\rho_{v} = \frac{H^{'}}{H}$ are purely related to the camera tilt angle $\theta$ and the camera height $h$.

To derive relation between the $\rho_{h}$, $\rho_{v}$ and the camera parameters, we simplify the rectangular area with width $W$ and height $H$ on the ground plane to a square area with a unit length, therefore $W=H=1$. Let us align the square at the corner as indicated as in Fig. \ref{fig:camera_setup_ratio}. The corner points on the unit square are $\bold{P}_{o}=\left[0, 0, 1 \right]^T$, $\bold{P}_{x}=\left[1, 0, 1 \right]^T$, $\bold{P}_{y}=\left[0, 1, 1 \right]^T$ and $\bold{P}_{xy}=\left[1, 1, 1 \right]^T$. Their corresponding pixel points are $\bold{p}_{o}$, $\bold{p}_{x}$, $\bold{p}_{y}$ and $\bold{p}_{xy}$, respectively:
\begin{align}
\centering
s\bold{p}_{o} &= \bold{K}\bold{t}\\ \nonumber
s\bold{p}_{x} &= \bold{K}\left[\bold{r}_{1}~\bold{t}\right]\\  \nonumber
s\bold{p}_{y} &= \bold{K}\left[\bold{r}_{2}~\bold{t}\right]\\  \nonumber
s\bold{p}_{xy} &= \bold{K}\left[\bold{r}_{1}~\bold{r}_{2}~\bold{t}\right],
\label{eq:projection_corners}
\end{align}

The horizontal ratio $\rho_h$ and vertical ratio $\rho_v$ is related to the camera model as following:
\begin{align}
\rho_h &=\frac{\left\| \bold{p}_{x} - \bold{p}_{o} \right\|}{\left\| \bold{p}_{xy} - \bold{p}_{y} \right\|}\\
\rho_v &=\left\| \bold{p}_{y} - \bold{p}_{o} \right\|.
\end{align}
Since pixels $\bold{p}_{o}$, $\bold{p}_{x}$, $\bold{p}_{y}$ and $\bold{p}_{xy}$ are only related the tilt angle $\theta$ and camera height $h$, $\rho_h$ and $\rho_v$ are therefore only related to these two parameters. 

\section{Further evaluation results}

In the main paper, due to restrictions in space, we solely reported the results in terms of F1-score, in this supplementary material, we provide the complete results, including the Precision and Recall values.  
Table \ref{table:supp_homography} corresponds to the results given in Table 1 of the main manuscript to compare the performance achieved using our proposed method for computing Homography versus the automatic approach proposed in state-of-the-art. 

Tables \ref{table:supp-leg}, \ref{table:supp-arm}, \ref{table:supp-torso}, \ref{table:supp-bbx} provides the Precision and Recall values corresponding to all experiments reported in Table 2 of the main manuscript, intending to investigate `Leg length', `Arm length', `Torso length', and the entire body in form of `BBX height' as the metric reference choice, using a fixed homography.

\begin{table*}[ht!]
	\centering
	\caption{Investigating the estimation of Homography matrix $\mathbf{H}$}
	\footnotesize
	\begin{tabular}{l|l|c c|c c c||c c c||c c c|}
		\cline{2-13}
		& \multicolumn{6}{c||}{Proposed VSD - Grid search}  & \multicolumn{3}{c||}{AutoRect $\mathbf{H}$} & \multicolumn{3}{c|}{Monoloco} \\ \hline
		\multicolumn{1}{|l|}{Dataset}                         & Seq. & $\rho_h$   & $\rho_v$   & Precision & Recall & F1-score & Precision  & Recall & F1-Score& Precision  & Recall & F1-Score \\ \hline \hline
		\multicolumn{1}{|l|}{\multirow{4}{*}{EPFL-mpv}}       & C0   & 0.7 & 0.5 &     78.26      &    78.26    &   77.90       &    70.04        &    81.05    &      73.47   & 70.56 & 78.42 & 73.46\\  
		\multicolumn{1}{|l|}{}                                & C1   & 0.5 & 0.6 &     74.45      &    77.03    &   75.39      &      51.27      &    87.72    &    61.17      & 68.54 & 73.87 & 70.19 \\  
		\multicolumn{1}{|l|}{}                                & C2   & 0.6 & 0.6 &     77.13      &    81.20    &   78.67      &    74.27        &    85.74    &    74.14      & 74.78 & 81.10 & 77.20\\  
		\multicolumn{1}{|l|}{}                                & C3   & 0.5 & 0.6 &     74.27      &    78.47    &   75.86      &      51.45      &    75.95    &    58.36      & 70.72 & 76.15 & 72.60\\ \hline \hline
		\multicolumn{1}{|l|}{\multirow{7}{*}{EPFL-wildtrack}} & C1   & 0.8 & 0.7 &    87.97       &    86.83    &   86.31       &       59.81     &    71.38    &     61.80     & 58.25 & 92.22 & 70.07\\  
		\multicolumn{1}{|l|}{}                                & C2   & 0.8 & 0.6 &    69.01       &    84.95    &   85.57       &     63.99       &    54.59    &     57.27     & 56.12 & 92.73 & 68.31\\  
		\multicolumn{1}{|l|}{}                                & C3   & 0.8 & 0.7 &    90.58       &    87.91    &   87.96       &      53.86      &     40.93   &    45.21      & 54.87 & 90.66 & 66.73 \\  
		\multicolumn{1}{|l|}{}                                & C4   & 0.6 & 0.8 &    89.35       &    84.37    &   85.54       &     42.84       &    29.12    &   35.06       & 53.84 & 85.97 & 64.29\\  
		\multicolumn{1}{|l|}{}                                & C5   & 0.8 & 0.8 &    59.86       &    91.38    &   69.91       &    39.56        &    81.10    &     50.99     & 40.78 & 91.93 & 54.88\\  
		\multicolumn{1}{|l|}{}                                & C6   & 0.8 & 0.8 &    60.92       &    76.06    &   65.27       &       49.97     &   36.58     &   39.54       & 36.25 & 89.31 & 49.64\\  
		\multicolumn{1}{|l|}{}                                & C7   & 0.5 & 0.7 &    88.17       &    87.73    &   86.96       &     41.37       &    94.23    &      55.63    & 57.57 & 88.17 & 68.44 \\ \hline \hline
		\multicolumn{1}{|l|}{OxTown}                          & -    & 0.5 & 0.8 &    82.98       &    82.30    &   81.04       &   37.38         &     95.16   &     51.78     & 42.46 & 83.78 & 54.57\\ \hline
	\end{tabular}
	\label{table:supp_homography}
\end{table*}

%%% LEG
\begin{table}[]
	\centering
	\caption{Investigating `Leg' body part as the metric reference choice, using the fixed homography}
	\footnotesize
	\begin{tabular}{llcc|ccc|}
		\cline{5-7}
		&                           &                          &     & \multicolumn{3}{c|}{Body parts - Leg length} \\ \hline
		\multicolumn{1}{|l|}{Dataset}                         & \multicolumn{1}{l|}{Seq.} & \multicolumn{1}{l}{$\rho_h$}   & $\rho_v$   & Precision      & Recall      & F1-score      \\ \hline \hline
		\multicolumn{1}{|l|}{\multirow{4}{*}{EPFL-mpv}}       & \multicolumn{1}{l|}{C0}   & \multicolumn{1}{l}{0.6} & 0.5 &    74.91       &   80.67     &   77.14       \\ 
		\multicolumn{1}{|l|}{}                                & \multicolumn{1}{l|}{C1}   & \multicolumn{1}{l}{0.5} & 0.6 &    72.63       &   77.97     &   74.80       \\
		\multicolumn{1}{|l|}{}                                & \multicolumn{1}{l|}{C2}   & \multicolumn{1}{l}{0.8} & 0.5 &    77.30       &   80.45     &   78.43       \\ 
		\multicolumn{1}{|l|}{}                                & \multicolumn{1}{l|}{C3}   & \multicolumn{1}{l}{0.8} & 0.5 &    73.31       &   74.54     &   73.41       \\ \hline \hline
		\multicolumn{1}{|l|}{\multirow{7}{*}{EPFL-WT}} & \multicolumn{1}{l|}{C1}   & \multicolumn{1}{l}{0.5} & 0.8 &    76.69       &   95.12     &   83.93       \\ 
		\multicolumn{1}{|l|}{}                                & \multicolumn{1}{l|}{C2}   & \multicolumn{1}{l}{0.7} & 0.5 &    72.12       &   96.12     &   83.93       \\
		\multicolumn{1}{|l|}{}                                & \multicolumn{1}{l|}{C3}   & \multicolumn{1}{l}{0.5} & 0.8 &    76.28       &   95.56     &   83.57       \\
		\multicolumn{1}{|l|}{}                                & \multicolumn{1}{l|}{C4}   & \multicolumn{1}{l}{0.6} & 0.8 &    86.94       &   87.69     &   85.95       \\
		\multicolumn{1}{|l|}{}                                & \multicolumn{1}{l|}{C5}   & \multicolumn{1}{l}{0.8} & 0.8 &    56.89       &   93.13     &   68.40       \\
		\multicolumn{1}{|l|}{}                                & \multicolumn{1}{l|}{C6}   & \multicolumn{1}{l}{0.8} & 0.7 &    51.09       &   89.08     &   62.98       \\
		\multicolumn{1}{|l|}{}                                & \multicolumn{1}{l|}{C7}   & \multicolumn{1}{l}{0.6} & 0.8 &    94.10       &   82.85     &   87.20       \\ \hline \hline
		\multicolumn{1}{|l|}{OxTown}                          & \multicolumn{1}{l|}{-}    & \multicolumn{1}{l}{0.5} & 0.8 &    80.61       &   87.62     &   82.59       \\ \hline
	\end{tabular}
	\label{table:supp-leg}
\end{table}

%%% ARM
\begin{table}[]
	\centering
	\caption{Investigating `Arm' body part as the metric reference choice, using the fixed homography}
	\footnotesize
	\begin{tabular}{llcc|ccc|}
		\cline{5-7}
		&                           &                          &     & \multicolumn{3}{c|}{Body parts - Arm length} \\ \hline
		\multicolumn{1}{|l|}{Dataset}                         & \multicolumn{1}{l|}{Seq.} & \multicolumn{1}{l}{$\rho_h$}   & $\rho_v$   & Precision      & Recall      & F1-score      \\ \hline \hline
		\multicolumn{1}{|l|}{\multirow{4}{*}{EPFL-mpv}}       & \multicolumn{1}{l|}{C0}   & \multicolumn{1}{l}{0.6} & 0.5 &    75.61       &   74.88     &   74.68       \\ 
		\multicolumn{1}{|l|}{}                                & \multicolumn{1}{l|}{C1}   & \multicolumn{1}{l}{0.5} & 0.6 &    72.09       &   71.26     &   71.19       \\
		\multicolumn{1}{|l|}{}                                & \multicolumn{1}{l|}{C2}   & \multicolumn{1}{l}{0.8} & 0.5 &    72.30       &   69.05     &   70.02       \\ 
		\multicolumn{1}{|l|}{}                                & \multicolumn{1}{l|}{C3}   & \multicolumn{1}{l}{0.8} & 0.5 &    69.74       &   64.61     &   66.38       \\ \hline \hline
		\multicolumn{1}{|l|}{\multirow{7}{*}{EPFL-WT}} & \multicolumn{1}{l|}{C1}   & \multicolumn{1}{l}{0.5} & 0.8 &    84.62       &   81.17     &   81.46       \\ 
		\multicolumn{1}{|l|}{}                                & \multicolumn{1}{l|}{C2}   & \multicolumn{1}{l}{0.7} & 0.5 &    81.36       &   82.80     &   80.47       \\
		\multicolumn{1}{|l|}{}                                & \multicolumn{1}{l|}{C3}   & \multicolumn{1}{l}{0.5} & 0.8 &    87.16       &   79.28     &   81.00       \\
		\multicolumn{1}{|l|}{}                                & \multicolumn{1}{l|}{C4}   & \multicolumn{1}{l}{0.6} & 0.8 &    86.87       &   66.59     &   73.30       \\
		\multicolumn{1}{|l|}{}                                & \multicolumn{1}{l|}{C5}   & \multicolumn{1}{l}{0.8} & 0.8 &    66.14       &   86.28     &   72.58       \\
		\multicolumn{1}{|l|}{}                                & \multicolumn{1}{l|}{C6}   & \multicolumn{1}{l}{0.8} & 0.7 &    59.61       &   73.57     &   63.53       \\
		\multicolumn{1}{|l|}{}                                & \multicolumn{1}{l|}{C7}   & \multicolumn{1}{l}{0.6} & 0.8 &    91.99       &   68.44     &   76.75       \\ \hline \hline
		\multicolumn{1}{|l|}{OxTown}                          & \multicolumn{1}{l|}{-}    & \multicolumn{1}{l}{0.5} & 0.8 &    82.66       &   69.01     &   73.03       \\ \hline
	\end{tabular}
	\label{table:supp-arm}
\end{table}

%%% TORSO
\begin{table}[]
	\centering
	\caption{Investigating `Torso' body part as the metric reference choice, using the fixed homography}
	\footnotesize
	\begin{tabular}{llcc|ccc|}
		\cline{5-7}
		&                           &                          &     & \multicolumn{3}{c|}{Body parts - Torso length} \\ \hline
		\multicolumn{1}{|l|}{Dataset}                         & \multicolumn{1}{l|}{Seq.} & \multicolumn{1}{l}{$\rho_h$}   & $\rho_v$   & Precision      & Recall      & F1-score      \\ \hline \hline
		\multicolumn{1}{|l|}{\multirow{4}{*}{EPFL-mpv}}       & \multicolumn{1}{l|}{C0}   & \multicolumn{1}{l}{0.6} & 0.5 &    76.44       &   79.88     &   77.64       \\ 
		\multicolumn{1}{|l|}{}                                & \multicolumn{1}{l|}{C1}   & \multicolumn{1}{l}{0.5} & 0.6 &    74.43       &   77.03     &   75.38       \\
		\multicolumn{1}{|l|}{}                                & \multicolumn{1}{l|}{C2}   & \multicolumn{1}{l}{0.8} & 0.5 &    77.11       &   77.88     &   77.12       \\ 
		\multicolumn{1}{|l|}{}                                & \multicolumn{1}{l|}{C3}   & \multicolumn{1}{l}{0.8} & 0.5 &    72.24       &   71.22     &   71.19       \\ \hline \hline
		\multicolumn{1}{|l|}{\multirow{7}{*}{EPFL-WT}} & \multicolumn{1}{l|}{C1}   & \multicolumn{1}{l}{0.5} & 0.8 &    79.39       &   91.80     &   84.10       \\ 
		\multicolumn{1}{|l|}{}                                & \multicolumn{1}{l|}{C2}   & \multicolumn{1}{l}{0.7} & 0.5 &    76.79       &   92.38     &   82.64       \\
		\multicolumn{1}{|l|}{}                                & \multicolumn{1}{l|}{C3}   & \multicolumn{1}{l}{0.5} & 0.8 &    82.06       &   91.89     &   85.37       \\
		\multicolumn{1}{|l|}{}                                & \multicolumn{1}{l|}{C4}   & \multicolumn{1}{l}{0.6} & 0.8 &    89.58       &   86.16     &   86.68       \\
		\multicolumn{1}{|l|}{}                                & \multicolumn{1}{l|}{C5}   & \multicolumn{1}{l}{0.8} & 0.8 &    59.47       &   91.82     &   69.82       \\
		\multicolumn{1}{|l|}{}                                & \multicolumn{1}{l|}{C6}   & \multicolumn{1}{l}{0.8} & 0.7 &    53.86       &   84.63     &   63.72       \\
		\multicolumn{1}{|l|}{}                                & \multicolumn{1}{l|}{C7}   & \multicolumn{1}{l}{0.6} & 0.8 &    93.07       &   78.36     &   84.03       \\ \hline \hline
		\multicolumn{1}{|l|}{OxTown}                          & \multicolumn{1}{l|}{-}    & \multicolumn{1}{l}{0.5} & 0.8 &    82.98       &   82.30     &   81.04       \\ \hline
	\end{tabular}
	\label{table:supp-torso}
\end{table}

%%% BBX
\begin{table}[]
	\centering
	\caption{Investigating `BBX height' as the metric reference choice, using the fixed homography}
	\footnotesize
	\begin{tabular}{llcc|ccc|}
		\cline{5-7}
		&                           &                          &     & \multicolumn{3}{c|}{Body parts - BBX height} \\ \hline
		\multicolumn{1}{|l|}{Dataset}                         & \multicolumn{1}{l|}{Seq.} & \multicolumn{1}{l}{$\rho_h$}   & $\rho_v$   & Precision      & Recall      & F1-score      \\ \hline \hline
		\multicolumn{1}{|l|}{\multirow{4}{*}{EPFL-mpv}}       & \multicolumn{1}{l|}{C0}   & \multicolumn{1}{l}{0.6} & 0.5 &    73.44       &   81.59     &   76.71       \\ 
		\multicolumn{1}{|l|}{}                                & \multicolumn{1}{l|}{C1}   & \multicolumn{1}{l}{0.5} & 0.6 &    70.85       &   78.39     &   73.90       \\
		\multicolumn{1}{|l|}{}                                & \multicolumn{1}{l|}{C2}   & \multicolumn{1}{l}{0.8} & 0.5 &    76.63       &   81.65     &   78.63       \\ 
		\multicolumn{1}{|l|}{}                                & \multicolumn{1}{l|}{C3}   & \multicolumn{1}{l}{0.8} & 0.5 &    73.38       &   77.24     &   74.69       \\ \hline \hline
		\multicolumn{1}{|l|}{\multirow{7}{*}{EPFL-WT}} & \multicolumn{1}{l|}{C1}   & \multicolumn{1}{l}{0.5} & 0.8 &    68.05       &   97.77     &   79.18       \\ 
		\multicolumn{1}{|l|}{}                                & \multicolumn{1}{l|}{C2}   & \multicolumn{1}{l}{0.7} & 0.5 &    71.97       &   96.56     &   81.18       \\
		\multicolumn{1}{|l|}{}                                & \multicolumn{1}{l|}{C3}   & \multicolumn{1}{l}{0.5} & 0.8 &    68.10       &   98.34     &   79.13       \\
		\multicolumn{1}{|l|}{}                                & \multicolumn{1}{l|}{C4}   & \multicolumn{1}{l}{0.6} & 0.8 &    74.15       &   94.55     &   81.52       \\
		\multicolumn{1}{|l|}{}                                & \multicolumn{1}{l|}{C5}   & \multicolumn{1}{l}{0.8} & 0.8 &    50.64       &   96.81     &   64.85       \\
		\multicolumn{1}{|l|}{}                                & \multicolumn{1}{l|}{C6}   & \multicolumn{1}{l}{0.8} & 0.7 &    54.26       &   93.09     &   59.73       \\
		\multicolumn{1}{|l|}{}                                & \multicolumn{1}{l|}{C7}   & \multicolumn{1}{l}{0.6} & 0.8 &    87.93       &   87.16     &   86.62       \\ \hline \hline
		\multicolumn{1}{|l|}{OxTown}                          & \multicolumn{1}{l|}{-}    & \multicolumn{1}{l}{0.5} & 0.8 &    63.18       &   90.12     &   72.38       \\ \hline
	\end{tabular}
	\label{table:supp-bbx}
\end{table}

\section{Social Distancing Violations Heatmaps}
In this section, we demonstrate the heatmap of social distancing violations over a short clip extracted from each dataset, as well as the footstep of the people on the ground shown in green dots. The correspondence of the density of footsteps and red area in the heatmap for all the examples can be appreciated.

\begin{figure*}
	\centering
	\includegraphics[width=2\columnwidth]{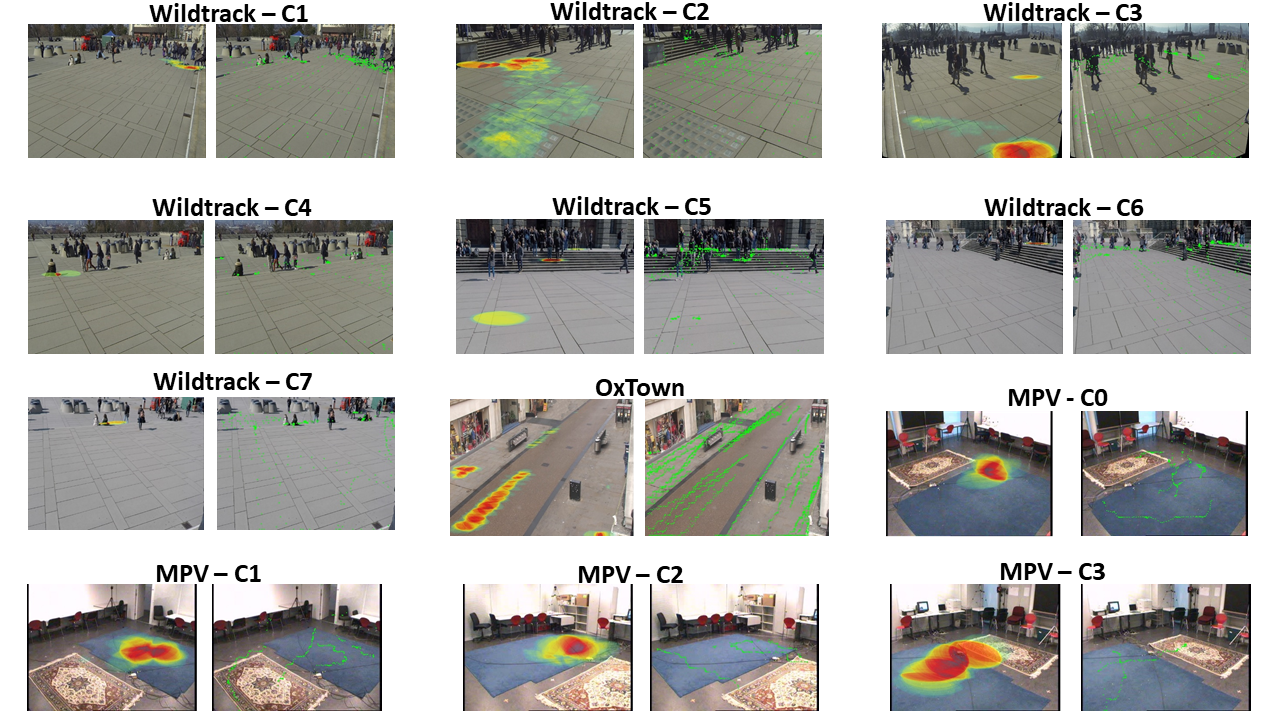}
	\caption{Image samples from all the sequences of the three datasets with the heatmap of social distancing violations overlaid.}
	\label{fig:dataset-heatmaps}
\end{figure*}

\section{Body-Part Joints}

In Table \ref{table:supp-joints}, we list the index of joints we used for defining a body part. This indexing indeed corresponds to the OpenPose 25 joint model output and might need adaption if any other pose detector is used.
\begin{table}[]
	\centering
	\caption{Body part joint index correspondence}
	\begin{tabular}{l c}
		\hline 
		Body part & Corresponding joints \\ \hline \hline
		Right arm & 5,6,7                \\ 
		Left arm  & 2,3,4                \\ 
		Right leg & 12,13,14,19          \\ 
		Left leg  & 9,10,11,22           \\ 
		Torso     & 1,8                  \\ \hline \hline
	\end{tabular}
	\label{table:supp-joints}
\end{table}

\end{document}